\documentclass[]{fairmeta}

\usepackage[table,dvipsnames]{xcolor}

\usepackage{microtype}
\usepackage{csquotes}
\usepackage{xspace}
\usepackage{enumitem}
\usepackage{nicefrac}
\usepackage{afterpage}
\usepackage{etoolbox}

\usepackage{amsmath}
\usepackage{amssymb}
\usepackage{mathtools}
\usepackage{amsthm}

\usepackage{graphicx}
\usepackage{wrapfig}
\usepackage{adjustbox}
\usepackage{booktabs}
\usepackage{colortbl}
\usepackage{hhline}
\usepackage{multirow}
\usepackage{threeparttable}
\usepackage{caption}

\usepackage{fontawesome}
\usepackage{pgfplots}
\pgfplotsset{compat=1.18}
\usepgfplotslibrary{groupplots}

\definecolor{GroupHeaderGray}{gray}{0.95}
\definecolor{RowLightBlue}{HTML}{E8F4FF}
\definecolor{baselinecolor}{gray}{.9}
\definecolor{tabbeige}{HTML}{F3EAD1}
\definecolor{tabblue}{HTML}{DDE7FF}
\definecolor{tabgreen}{HTML}{DCE9A6}
\definecolor{fbApp}{HTML}{c8e7fa}
\definecolor{fbPurple3}{HTML}{f0ebf5}

\definecolor{deemph}{gray}{0.6}
\definecolor{citecolor}{HTML}{0071BC}
\definecolor{linkcolor}{HTML}{ED1C24}

\definecolor{darkbg}{RGB}{30,30,30}

\definecolor{bestgreen}{RGB}{183,223,185}
\definecolor{secondgreen}{RGB}{223,240,216}

\newbool{darkmode}
\boolfalse{darkmode}

\newcommand{\adaptivetext}[1]{%
    \ifbool{darkmode}{\textcolor{black}{#1}}{#1}%
}

\newcommand{\adaptivebold}[1]{%
    \ifbool{darkmode}{\textcolor{black}{\textbf{#1}}}{\textbf{#1}}%
}

\newcolumntype{x}[1]{>{\centering\arraybackslash}p{#1pt}}
\newcolumntype{y}[1]{>{\raggedright\arraybackslash}p{#1pt}}
\newcolumntype{z}[1]{>{\raggedleft\arraybackslash}p{#1pt}}

\newlength\savewidth

\newcommand{\best}[1]{\cellcolor{bestgreen}\textbf{#1}}
\newcommand{\second}[1]{\cellcolor{secondgreen}#1}

\newcommand{\finding}[2]{%
    \begin{tcolorbox}[
        colback=white!90!gray,
        colframe=teal!60!black,
        arc=5pt,
        boxsep=5pt,
        left=10pt,
        right=10pt,
        top=2pt,
        bottom=2pt,
        boxrule=0.8pt,
        drop shadow=gray!50!white,
        enhanced jigsaw,
    ]
    #2
    \end{tcolorbox}%
}

\renewcommand{\paragraph}[1]{\vspace{1.25mm}\noindent\textbf{#1}}

\makeatletter
\newcommand{\ssymbol}[1]{$^{\@fnsymbol{#1}}$}
\makeatother

\newcommand{\app}{\raise.17ex\hbox{$\scriptstyle\sim$}}

\newcommand{\newrelease}[1]{\colorbox{bestgreen}{\strut\,\textbf{#1}\,}}

\definecolor{main}{HTML}{2A6175}
\definecolor{Secondary}{HTML}{7FB0A3}
\newcommand{\mainnum}[1]{\textcolor{main}{#1}}

\newcommand{\baselinenum}[1]{\textcolor{black!55}{#1}}
\definecolor{MolmoPink}{HTML}{F7A0C9}

\usepackage{booktabs}
\usepackage{multirow}
\usepackage{array}
\usepackage{xcolor}
\usepackage{colortbl}
\usepackage{adjustbox}

\newcommand{\mainvideobenchmarktable}{
\begin{table*}[t]
\centering
\setlength{\tabcolsep}{6pt}
\renewcommand{\arraystretch}{1.3}
\caption{\textbf{Video understanding benchmarks for 8B-class MLLMs.} \textbf{LLaVA-OneVision-2-8B} leads the 18-task video understanding average (62.5) and the 4-task RVOS tracking average (48.0, J\&F). We use \colorbox{bestgreen}{\phantom{xx}} and \colorbox{secondgreen}{\phantom{xx}} to denote the best and second-best results.}
\label{tab:video_main}
\adjustbox{max width=\textwidth}{
\begin{tabular}{llcccccc}
\toprule
\textbf{Group} & \textbf{Benchmark}
& \textbf{LLaVA-OV-2}
& \textbf{Qwen3-VL}
& \textbf{Keye-VL-1.5}
& \textbf{InternVL-3.5}
& \textbf{PLM}
& \textbf{LLaVA-OV-1.5} \\
\midrule

\multirow{13}{*}{\textbf{\shortstack[l]{Video\\QA}}}
& MV-Bench               & 66.2          & 69.0          & 56.9          & \second{72.1} & \best{77.1}   & 51.2 \\
& NextQA                 & 82.5          & \second{83.4} & 75.8          & 82.0          & \best{84.1}   & 73.7 \\
& TempCompass-MC         & \second{74.5} & 74.3          & \best{75.5}   & 70.4          & 72.7          & 57.5 \\
& VideoMME               & \second{71.9} & 71.4          & \best{73.0}   & 65.9          & 60.5          & 61.1 \\
& VideoMME (w/ sub.)     & \best{76.3}   & 75.6          & \second{76.2} & 68.6          & 65.6          & 65.5 \\
& VideoMME-v2-64         & \best{19.9}   & \second{18.2} & 14.1          & 14.6          & 8.7           & 9.1  \\
& MLVU-dev               & \second{76.6} & \best{78.1}   & 75.0          & 71.0          & 66.4          & 62.1 \\
& LVBench                & \second{55.5} & \best{58.0}   & 42.8          & 46.7          & 44.5          & 40.1 \\
& LongVideoBench         & \second{66.9} & \best{68.0}   & 66.0          & 62.4          & 59.6          & 56.2 \\
& MMVU-val               & 56.2          & 58.7          & \best{68.3}   & \second{60.2} & 43.3          & 50.1 \\
& VideoEval-Pro          & \best{61.5}   & \second{59.2} & 54.9          & 50.1          & 47.2          & 44.8 \\
& MMOU                   & \second{39.5} & \best{40.6}   & 35.3          & 36.1          & 26.2          & 30.7 \\
& JumpScore & \best{74.9} & 30.1    & \second{39.6} & 11.0          & 13.1          & 2.1  \\
\midrule

\multirow{3}{*}{\textbf{\shortstack[l]{Temp.\\Ground.}}}
& t/Charades             & \best{53.5}   & \second{48.3} & 45.4          & 27.8          & 34.5          & 15.6 \\
& t/ActivityNet          & \best{53.8}   & \second{46.8} & 41.3          & 31.3          & 7.6           & 17.7 \\
& t/QVHighlights         & \best{66.4}   & \second{59.4} & 55.5          & 31.3          & 4.2           & 21.0 \\
\midrule

\multirow{2}{*}{\textbf{\shortstack[l]{Vis.\\Spatial}}}
& VSI-Bench              & \best{70.9}   & \second{59.1} & 36.4          & 56.0          & 27.9          & 30.2 \\
& ReVSI                  & \best{57.6}   & \second{48.9} & 32.4          & 47.9          & 30.7          & 33.5 \\
\midrule
\multicolumn{2}{l}{\textbf{Video Average (18 tasks)}}
& \best{62.5} & \second{58.2} & 53.6 & 50.3 & 43.0 & 40.1 \\
\midrule

\multirow{4}{*}{\textbf{\shortstack[l]{Tracking\\(J\&F)}}}
& DAVIS                  & \best{58.7}   & \second{41.3} & 5.8           & 4.7           & 2.0           & 4.1  \\
& MeViS$_U$              & \best{45.7}   & \second{28.4} & 7.2           & 7.5           & 7.6           & 6.1  \\
& ReVOS-ref              & \best{58.2}   & \second{37.8} & 10.7          & 10.2          & 8.5           & 13.0 \\
& ReVOS-reason           & \best{29.2}   & \second{21.9} & 9.6           & 9.2           & 10.2          & 9.7  \\
\midrule
\multicolumn{2}{l}{\textbf{Tracking Average (4 tasks, J\&F)}}
& \best{48.0} & \second{32.4} & 8.3 & 7.9 & 7.1 & 8.2 \\
\bottomrule
\end{tabular}}
\end{table*}
}

\newcommand{\mainspatialimagebenchmarktable}{
\begin{table*}[t]
\centering
\setlength{\tabcolsep}{6pt}
\renewcommand{\arraystretch}{1.3}
\caption{\textbf{Spatial reasoning, image and document benchmarks for 8B-class MLLMs.} \textbf{LLaVA-OneVision-2-8B} is most differentiated on spatial reasoning (notably \textbf{CrossPoint}, \textbf{TraceSpatial-3D}), while staying competitive on image / document understanding and leading V$^{*}$-Bench. We use \colorbox{bestgreen}{\phantom{xx}} and \colorbox{secondgreen}{\phantom{xx}} to denote the best and second-best results.}
\label{tab:spatial_main}
\adjustbox{max width=\textwidth}{
\begin{tabular}{llcccccc}
\toprule
\textbf{Group} & \textbf{Benchmark}
& \textbf{LLaVA-OV-2}
& \textbf{Qwen3-VL}
& \textbf{Keye-VL-1.5}
& \textbf{InternVL-3.5}
& \textbf{PLM}
& \textbf{LLaVA-OV-1.5} \\
\midrule

\multirow{11}{*}{\textbf{\shortstack[l]{Spatial\\Reasoning}}}
& CRPE            & \second{77.3} & \best{77.7}   & 75.2          & 75.0          & 77.0          & 74.8 \\
& MetaVQA         & \best{69.1}   & \second{68.7} & 59.2          & 65.7          & 45.4          & 67.1 \\
& ERQA            & \second{43.3} & 42.3          & 38.3          & 41.8          & \best{44.3}   & 41.5 \\
& CV-Bench 2D     & \best{82.6}   & \second{81.0} & 78.2          & 77.9          & 80.6          & 76.5 \\
& CV-Bench 3D     & \best{92.8}   & \second{92.3} & 82.0          & 86.3          & 82.4          & 82.9 \\
& CrossPoint      & \best{61.9}   & \second{26.9} & 20.2          & 20.2          & 15.7          & 15.9 \\
& EmbSpatial      & \best{78.1}   & \second{77.5} & 66.3          & 73.2          & 73.5          & 64.2 \\
& SAT             & \best{69.3}   & \best{69.3}   & \second{62.7} & 54.7          & 36.7          & 61.3 \\
& MMSI-Bench      & 29.6          & \second{31.0} & 26.7          & 28.1          & \best{31.4}   & 28.3 \\
& BLINK           & \second{63.5} & \best{65.1}   & 52.2          & 55.7          & 56.0          & 48.3 \\
& TraceSpatial-3D & \best{31.0}   & \second{8.0}  & 3.0           & 4.0           & 1.0           & 1.0  \\
\midrule

\multicolumn{2}{l}{\textbf{Spatial Average (11 tasks)}}
& \best{63.5} & \second{58.2} & 51.3 & 53.0 & 49.5 & 51.1 \\
\midrule

\multirow{11}{*}{\textbf{\shortstack[l]{Image\\\&\\Document}}}
& MMStar          & 64.8          & 62.9          & \best{73.6}   & 66.6          & 57.9          & \second{67.9} \\
& MMBench-en      & 85.7          & 84.9          & \best{88.5}   & \second{87.9} & 80.2          & 85.6 \\
& DocVQA          & 95.2          & \second{95.7} & 94.9          & 92.3          & 94.6          & \best{97.8} \\
& ChartQA         & 85.9          & 85.1          & 84.7          & \best{86.7}   & 85.5          & \second{86.5} \\
& InfoVQA         & 74.4          & \best{83.4}   & 76.9          & 79.1          & \second{80.0} & 79.1 \\
& OCRBench        & 78.2          & \second{84.7} & \best{84.8}   & 84.0          & 83.2          & 82.6 \\
& AI2D            & 84.3          & 83.6          & \second{86.0} & 84.0          & \best{92.7}   & 84.0 \\
& V$^{*}$-Bench   & \best{85.9}   & \second{85.3} & 78.0          & 81.7          & 71.2          & 77.5 \\
& CountBench      & 89.0          & \second{89.8} & 83.1          & 75.6          & \best{91.8}   & 87.8 \\
& Pixmo-Count     & \second{64.0} & 62.4          & 55.6          & 61.8          & \best{68.0}   & 63.1 \\
& RealWorldQA     & 69.7          & 69.4          & \second{69.8} & 63.1          & \best{72.7}   & 68.1 \\
\midrule

\multicolumn{2}{l}{\textbf{Image \& Document Average (11 tasks)}}
& 79.7 & \best{80.7} & 79.6 & 78.4 & 79.8 & \second{80.0} \\

\bottomrule
\end{tabular}}
\end{table*}
\vspace{-2mm}
}

\newcommand{\trackingbenchmarktable}{
\begin{table*}[h]
\centering
\setlength{\tabcolsep}{4pt}
\renewcommand{\arraystretch}{1.1}
\caption{\textbf{Referring video object segmentation and point-to-mask tracking.}
Evaluation on DAVIS, MeViS-U, REVOS-Referring, and REVOS-Reasoning with three metrics: F, HOTA, and J\&F.
For LLaVA-OneVision-2-8B, predicted ID-associated point tracks are converted into masks using SAM~2 point prompts before computing mask-level metrics.
The final Overall column reports the aggregate tracking score used by our evaluation protocol, rather than the unweighted mean of the four J\&F columns.
\textbf{Bold} indicates the best result in each column.}
\label{tab:tracking_main}
\adjustbox{max width=\textwidth}{
\begin{tabular}{l|ccc|ccc|ccc|ccc|c}
\toprule

& \multicolumn{3}{c|}{\textbf{DAVIS}}
& \multicolumn{3}{c|}{\textbf{MeViS-U}}
& \multicolumn{3}{c|}{\textbf{REVOS (Ref.)}}
& \multicolumn{3}{c|}{\textbf{REVOS (Reason)}}
& \textbf{Overall} \\

\textbf{Method}
& F & HOTA & J\&F
& F & HOTA & J\&F
& F & HOTA & J\&F
& F & HOTA & J\&F
&  \\

\midrule

InternVL3.5-8B
& 12.8 & 9.4 & 4.7
& 7.2 & 6.6 & 7.5
& 22.2 & 20.8 & 10.2
& 7.9 & 7.2 & 9.2
& 11.0 \\

LLaVA-OV-1.5
& 11.9 & 8.8 & 4.1
& 7.3 & 9.0 & 6.1
& 16.8 & 17.4 & 13.0
& 6.2 & 17.8 & 9.7
& 10.8 \\

Keye-VL-1.5
& 14.6 & 12.2 & 5.8
& 10.1 & 15.8 & 7.2
& 22.1 & 26.5 & 10.7
& 9.9 & 12.0 & 9.6
& 12.1 \\

PLM-8B
& 7.8 & 32.8 & 2.0
& 5.0 & 34.4 & 7.6
& 6.8 & 51.2 & 8.5
& 0.1 & 0.1 & 10.2
& 9.0 \\

Qwen3-VL-8B
& 39.7 & 36.6 & 41.3
& 29.9 & \textbf{34.5} & 28.4
& 40.7 & 41.5 & 37.8
& 24.7 & \textbf{32.4} & 21.9
& 30.8 \\

\midrule

\mainnum{\textbf{LLaVA-OV-2-8B}}
& \mainnum{\textbf{52.7}} & \mainnum{\textbf{44.7}} & \mainnum{\textbf{58.7}}
& \mainnum{\textbf{37.1}} & \mainnum{31.7} & \mainnum{\textbf{45.7}}
& \mainnum{\textbf{60.8}} & \mainnum{\textbf{56.1}} & \mainnum{\textbf{58.2}}
& \mainnum{\textbf{27.4}} & \mainnum{22.5} & \mainnum{\textbf{29.2}}
& \mainnum{\textbf{41.0}} \\

\bottomrule
\end{tabular}}
\end{table*}
}

\newcommand{\longvideocodectable}{
\begin{table*}[!t]
\centering
\setlength{\tabcolsep}{4.5pt}
\renewcommand{\arraystretch}{1.1}
\caption{\textbf{Extended long-video frame-budget sweep on LVBench, VideoMME-Long, MLVU-dev, and VideoEval-Pro.} Stream inputs hold parity or a small lead at low budgets and converge closely with Fix inputs at higher budgets.}
\label{tab:longvideo_extended}
\adjustbox{max width=\textwidth}{
\begin{tabular}{lcccccccc}
\toprule
\multirow{2}{*}{\textbf{Token Budget}} 
& \multicolumn{2}{c}{\textbf{LVBench}} 
& \multicolumn{2}{c}{\textbf{VideoMME-L (w/ sub)}} 
& \multicolumn{2}{c}{\textbf{MLVU-dev}} 
& \multicolumn{2}{c}{\textbf{VideoEval-Pro}} \\
\cmidrule(lr){2-3} 
\cmidrule(lr){4-5} 
\cmidrule(lr){6-7} 
\cmidrule(lr){8-9}
& \baselinenum{Fix} & \mainnum{Stream} 
& \baselinenum{Fix} & \mainnum{Stream} 
& \baselinenum{Fix} & \mainnum{Stream} 
& \baselinenum{Fix} & \mainnum{Stream} \\
\midrule
2k   
& \baselinenum{\textbf{40.0}} & \mainnum{38.5} 
& \baselinenum{55.1} & \mainnum{\textbf{56.6}} 
& \baselinenum{60.0} & \mainnum{\textbf{62.4}} 
& \baselinenum{45.0} & \mainnum{\textbf{46.4}} \\

4k  
& \baselinenum{40.4} & \mainnum{\textbf{40.9}} 
& \baselinenum{57.7} & \mainnum{\textbf{60.3}} 
& \baselinenum{62.6} & \mainnum{\textbf{65.0}} 
& \baselinenum{45.3} & \mainnum{\textbf{48.3}} \\

8k 
& \baselinenum{\textbf{43.4}} & \mainnum{42.9} 
& \baselinenum{\textbf{62.0}} & \mainnum{61.4} 
& \baselinenum{67.1} & \mainnum{\textbf{69.0}} 
& \baselinenum{48.1} & \mainnum{\textbf{51.6}} \\

16k 
& \baselinenum{46.7} & \mainnum{\textbf{47.1}} 
& \baselinenum{64.8} & \mainnum{\textbf{67.4}} 
& \baselinenum{70.2} & \mainnum{\textbf{71.3}} 
& \baselinenum{51.4} & \mainnum{\textbf{53.9}} \\

32k 
& \baselinenum{48.8} & \mainnum{\textbf{49.5}} 
& \baselinenum{\textbf{67.1}} & \mainnum{67.0} 
& \baselinenum{72.5} & \mainnum{\textbf{73.7}} 
& \baselinenum{\textbf{56.0}} & \mainnum{55.8} \\
\bottomrule
\end{tabular}}
\end{table*}
}

\title{LLaVA-OneVision-2: Towards Next-Generation Perceptual Intelligence}

\author{Glint Lab}
\author{AIM for Health Lab}
\author{MVP Lab}

\abstract{
\noindent We introduce \textbf{LLaVA-OneVision-2~(LLaVA-OV-2)}, the most capable vision-language model in the LLaVA-OneVision series to date, achieving superior performance across a broad range of multimodal benchmarks. The model builds on a native OneVision-Encoder and incorporates Windowed Attention for efficient local computation while maintaining native resolution. Its key advance is \textbf{codec-stream tokenization}: it treats compressed video as a continuous bit-cost stream, where bit-cost dynamics determine adaptive temporal groups, and motion-residual cues select salient spatial evidence into compact visual canvases. This allocation concentrates a limited token budget on event-bearing content, enabling more stable long-video token compression than fixed groups of pictures. A shared 3D RoPE further places codec canvases, sampled frames, and images in a unified spatiotemporal coordinate system. Furthermore, we build the LLaVA-OV-2 data and training stack around large-scale open supervision: approximately \(8\)M re-captioned video samples for pretraining, a \(4\)M-sample spatial corpus for fine-tuning. We also introduce \textbf{JumpScore}, a temporal-localization benchmark targeting fine-grained grounding in high-frequency, densely repeated motion, a regime underrepresented by existing video evaluations. A standout capability of LLaVA-OV-2 is its unified perception across video understanding, temporal grounding, spatial grounding, and manipulation-trace reasoning. On JumpScore, LLaVA-OneVision-2-8B reaches \(74.9\) JumpScore mAP, surpassing Qwen3-VL-8B (\(30.1\)) by \(+44.8\) points; under matched visual-token budgets on the same benchmark, codec-stream inputs improve temporal grounding over frame sampling by \(+9.7\) points. Across standard benchmarks, LLaVA-OneVision-2-8B further outperforms Qwen3-VL-8B by \(+4.3\) average points on video tasks, \(+5.3\) on spatial tasks, and \(+15.6\) average J\&F on tracking tasks. Our code, data, and models are released as open-source resources.
}

\date{\today}
\metadata[Code]{\url{https://github.com/EvolvingLMMs-Lab/LLaVA-OneVision-2}}
\metadata[Data]{\url{https://huggingface.co/datasets/mvp-lab/LLaVA-OneVision-2-Data}}
\metadata[Model]{\url{https://huggingface.co/lmms-lab-encoder/LLaVA-OneVision-2-8B-Instruct}}

\begin{document}
\maketitle

\begin{figure*}[!t]
\centering
\includegraphics[width=\linewidth]{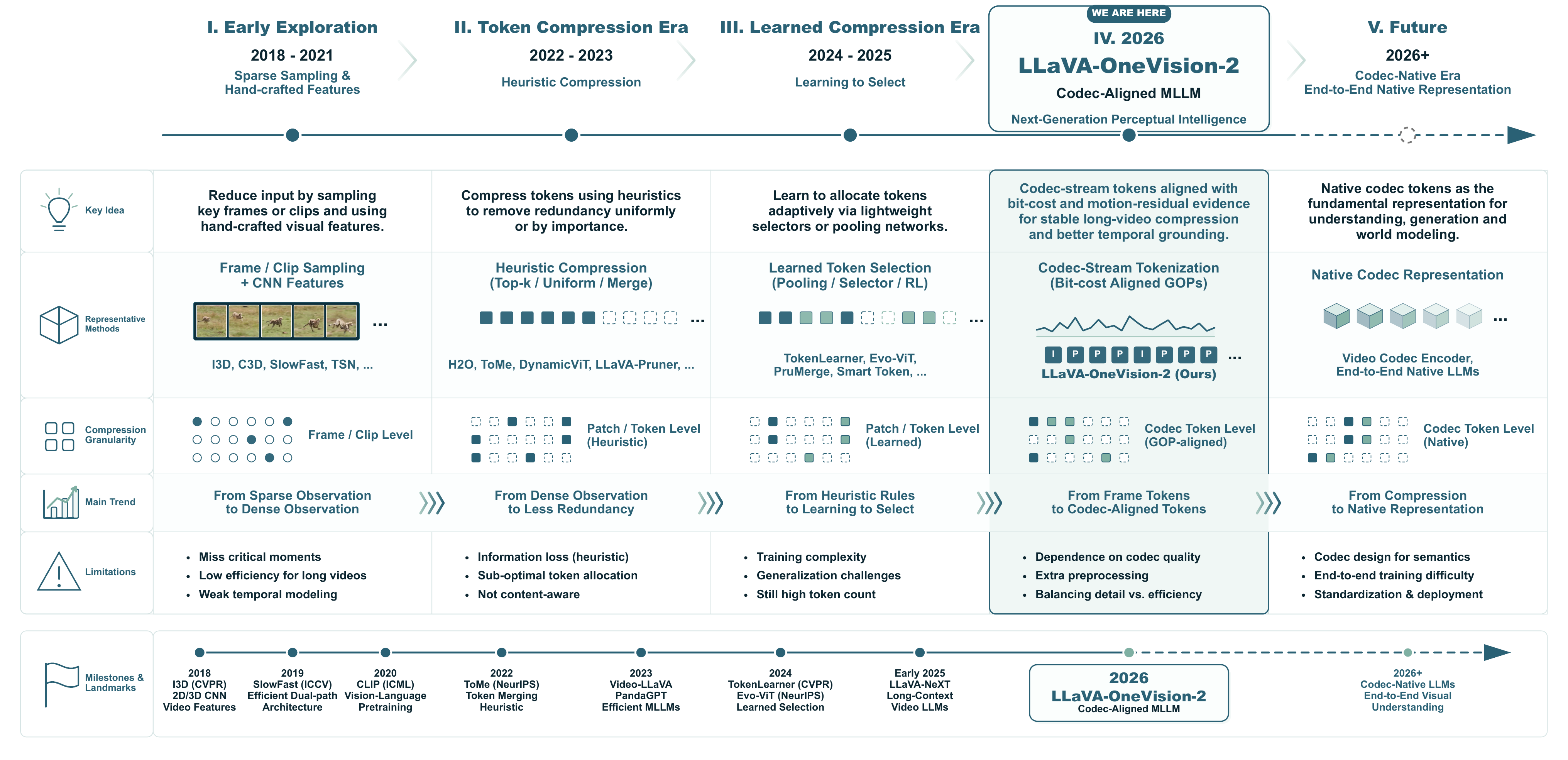}
\caption{\textbf{Roadmap of video understanding from token compression to codec-aligned perceptual intelligence.} The roadmap traces the evolution from early frame/clip sampling and hand-crafted visual features, to heuristic token compression, learned token selection, and the 2026 codec-aligned paradigm represented by LLaVA-OneVision-2.}
\label{fig:teaser}
\end{figure*}

\section{Introduction}
\label{sec:intro}

Recent open Large Vision-Language Models (LVLMs)~\citep{Qwen3-VL,bai2025qwen25vl,internvl3,an2025llava,Keye-VL-1.5,cambrian-s,zhang2026penguin,clark2026molmo2,liu2024nvila,zhang2025videollama3,zohar2024apollo,wang2025internvideo25,liu2024oryx,longvu} largely retain a frame-centric observation paradigm: \emph{Uniform frame sampling or Mixed-resolution frames}, which combines sparse high-resolution key frames with denser low-resolution context frames to satisfy a fixed token budget. Yet such designs still reduce video to a set of decoded frames, underrepresenting continuous spatial structure and motion dynamics while overlooking the predictive stream signals that make video uniquely informative. Video codecs such as H.264 and H.265/HEVC \textit{(High Efficiency Video Coding)} decompose video signals into spatially complete intra-coded frames (I-frames) that establish global context and predicted frames (P-frames) that encode inter-frame variations via motion compensation and residuals~\citep{sullivan2012overview}. The OneVision-Encoder (OV-Encoder)~\citep{ovencoder} is an early prototype along this path: it introduced codec patchification as a backbone-side primitive and showed that, under a fixed token budget, codec-selected I/P patches provide the language model with denser discriminative evidence than uniformly sampled frame patches.

In this paper, we argue that next-generation perceptual intelligence should move beyond uniformly observing frames toward selectively allocating evidence in predictive visual streams, where most pixels sustain contextual continuity and only sparse deviations encode discriminative semantic, spatial, and temporal structure.

We introduce \textbf{LLaVA-OneVision-2}, the most capable vision-language model in the LLaVA-OneVision series to date, achieving strong performance across a broad range of multimodal benchmarks. The model builds on a native dynamic-resolution OV-Encoder with codec patchification, and augments it with a codec-adaptive attention interface that combines spatial windowed attention for efficient local computation with group-visible masks while preserving native resolution. Its key advance is \textbf{codec-stream tokenization}: It treats compressed video as a continuous bit-cost stream. Bit-cost dynamics adaptively determine temporal group boundaries, while motion-residual cues condense salient spatial evidence into compact, merge-aligned visual canvases. Rather than allocating visual tokens by elapsed time or fixed frame slots, this stream-aware design makes token density follow the evolving bit-cost-residual profile of the compressed stream, densifying around perceptual transitions while thinning over predictable intervals, thereby enabling more stable long-video token compression than fixed Group of Pictures (GOPs). A shared 3D RoPE further places codec canvases, sampled frames, and images in a unified spatiotemporal coordinate system.

LLaVA-OneVision-2 is trained with a progressive four-stage recipe that scales supervision from image grounding to long-video and spatial reasoning. Stage 1 mixes \(\sim\!85\)M image--text samples with 4.2M 30s video captions at 30 frames; Stage 2 adds large-scale instruction data (\(\sim\!22\)M LLaVA-OneVision-1.5 and \(\sim\!24\)M FineVision samples) together with 2.7M 30--60s and 700K 60--180s video captions at 60/90 frames; Stage 3 extends to long-form video with video-instruction corpora and 350K 10--15min captions at 384 frames; and Stage 4 re-encodes long videos through the variable-length-GOP codec pipeline at 384/768 frames while adding LLaVA-OneVision-2-Spatial-4M for 2D/3D spatial supervision. Across all stages, codec-patchified videos, uniformly sampled videos, and image/tiled inputs are interleaved in training.

Existing video benchmarks underrepresent fine-grained grounding in high-frequency, densely repeated motion, where the challenge is not merely recognizing the event category but localizing the correct action instance among many visually similar cycles. We therefore introduce JumpScore, a temporal-localization benchmark designed to evaluate perceptual transition-level grounding. On JumpScore, LLaVA-OneVision-2-8B achieves \(74.9\) JumpScore mAP, substantially outperforming Qwen3-VL-8B (\(30.1\)) by \(+44.8\) points. Within the same benchmark and under matched visual-token budgets, codec-stream inputs improve temporal grounding over frame-sampling inputs by \(+9.7\) points. At the model level, LLaVA-OneVision-2-8B outperforms Qwen3-VL-8B by \(+4.3\) points on average across video benchmarks, \(+5.3\) points across spatial benchmarks, and \(+15.6\) average J\&F on tracking benchmarks. Our experiments have revealed: codec-stream inputs favor long-video tasks governed by coarse temporal structure, such as temporal grounding, event understanding, event ordering, and salient retrieval, by reallocating tokens to high-bit-cost intervals and high-residual regions. In contrast, frame sampling remains preferable for detail-sensitive queries, where decisive cues are static, fine-grained, spatially small, trajectory-specific, or boundary-level, because dense frame observations better preserve local texture, subtle appearance cues, and frame-to-frame continuity.

In summary, the main contributions of LLaVA-OneVision-2 are as follows:
\begin{enumerate}[leftmargin=1.4em,topsep=0.2em,itemsep=0.15em]
    \item LLaVA-OneVision-2 is a codec-aligned MLLM whose codec-stream tokenization treats video as a continuous bit-cost stream, aligning visual-token allocation with bit-cost dynamics and motion-residual evidence to enable stable long-video token compression.
    \item We scale training with approximately \(8\)M re-captioned video samples and a \(4\)M-sample 2D/3D spatial corpus, and introduce JumpScore, a temporal-localization benchmark for fine-grained video grounding in high-frequency, dense motion. Our code, data, and models are released.
    \item LLaVA-OV-2-8B delivers consistent gains over Qwen3-VL-8B, improving the average score by \(+4.3\) points across 18 video tasks, \(+5.3\) points across 11 spatial-reasoning tasks, and \(+15.6\) average J\&F across 4 tracking tasks. Codec-stream inputs further improve temporal grounding over frame sampling by \(+9.7\) points, and LLaVA-OV-2-8B reaches \(74.9\) JumpScore mAP against Qwen3-VL-8B's \(30.1\) (\(+44.8\) points).
\end{enumerate}
\section{Architecture}
\label{sec:architecture}

\begin{figure*}[t]
\centering
\includegraphics[width=\textwidth]{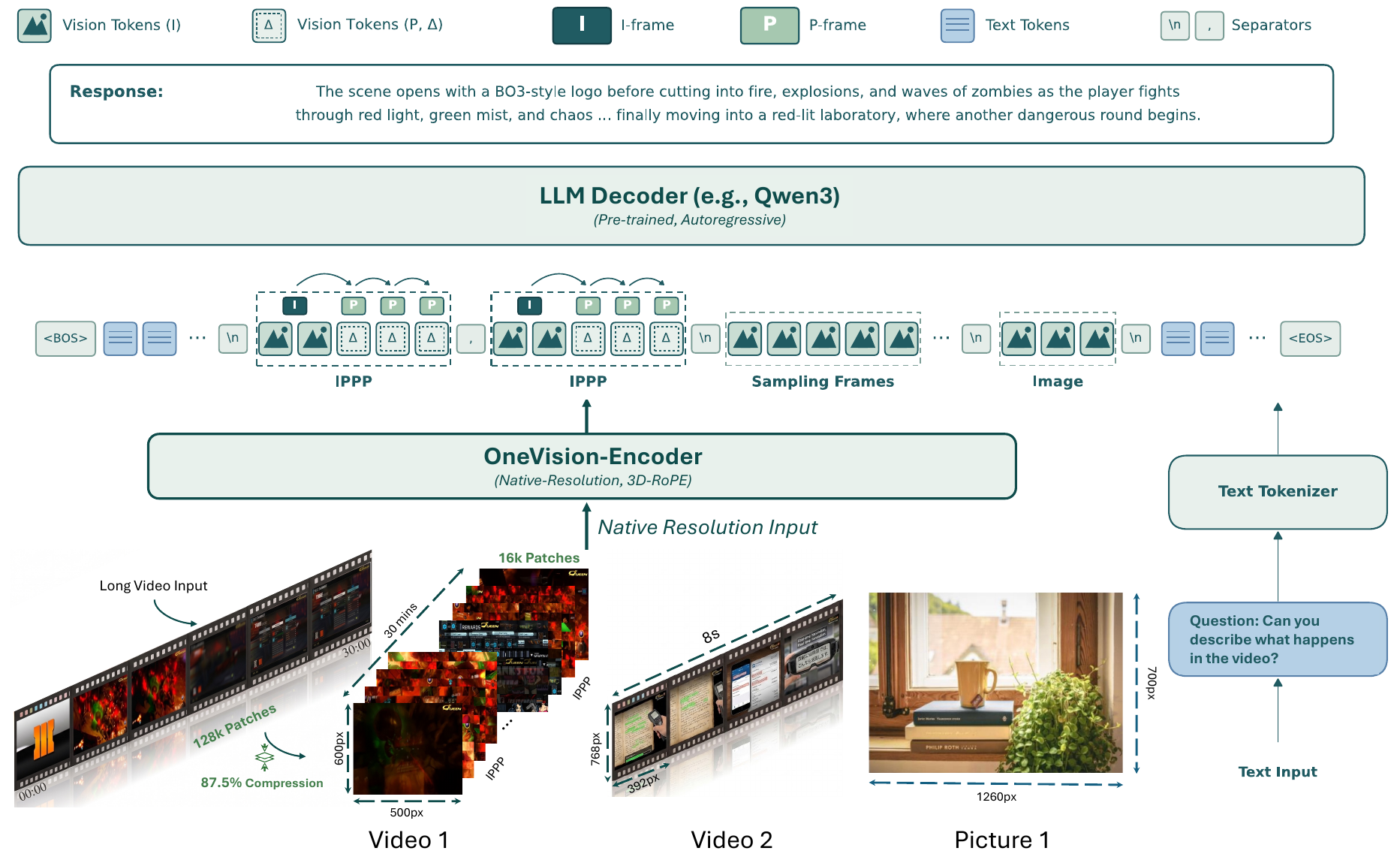}
\caption{\textbf{LLaVA-OneVision-2 architecture.}
The model unifies codec-stream videos, sampled-frame videos, and native-resolution images under a shared visual-token interface. Codec inputs are encoded as I/P visual canvases, sampled videos as frame-token sequences, and images as spatial visual tokens; all inputs are processed by the OneVision-Encoder. The resulting visual embeddings are combined with text tokens and decoded by a pre-trained autoregressive language model, allowing a single architecture to support video and image understanding.
}
\label{fig:arch}
\end{figure*}

This section describes the model-side design of LLaVA-OneVision-2, as illustrated in Figure~\ref{fig:arch}.
\S\ref{subsec:full_system} first gives the full multimodal stack, consisting of a Vision Encoder, a lightweight vision-language connector, and an autoregressive language model decoder.
\S\ref{sec:codec_pipeline} then focuses on the visual encoder interface: how sampled frames, codec-patchified videos, and static images are represented as visual canvases with token metadata, and group-visible attention masks.

\subsection{LLaVA-OneVision-2}
\label{subsec:full_system}

\paragraph{Vision Encoder.}
LLaVA-OneVision-2 adopts the OneVision-Encoder~\citep{ovencoder} as a shared backbone for sampled-frame videos, codec-stream videos, and static images, mapping all inputs into a unified visual-token interface with patch embeddings, 3D positional coordinates, and encoder-side group assignments. Shared 3D RoPE provides a common spatiotemporal coordinate system, while group-visible masks define token visibility: sampled-frame and IPPP-style inputs use fixed four-slot groups, static images use a degenerate single-temporal group, and codec-stream inputs use bit-cost-adaptive GOP ids to group tokens from the same variable-length GOP across P-canvases. Following native-resolution vision-transformer designs~\citep{dehghani2023navit,beyer2023flexivit,siglip2,bai2025qwen25vl}, spatial windowed attention is used in most visual layers for efficient native-resolution processing and remains orthogonal to the video-level grouping rule.

\paragraph{Vision-Language Connector.}
A lightweight two-layer MLP maps OneVision-Encoder representations into the language-model embedding space.  Because sampled-frame videos, IPPP-style windows, codec-derived I/P canvases, and static images share the same encoder-output format, the connector remains interface-invariant across input forms. Codec-stream processing therefore changes only the evidential structure presented to the visual encoder, while leaving the vision-language alignment interface unchanged.

\paragraph{Large Language Model.}
The projected visual tokens are paired with the text instruction and decoded by a shared Qwen3-8B autoregressive language model under the supervised next-token objective. No codec-specific adapter, reconstruction decoder, or language-side branch is introduced. Consequently, frame-sampled and codec-stream inputs differ only in evidence selection and attention-group assignment, whereas the encoder--connector--decoder pathway remains architecturally identical.

\subsection{Codec-stream Tokenization}
\label{sec:codec_pipeline}

\paragraph{Unified Visual-token Interface.}
For a video \(\mathcal{V}\), the codec front-end emits visual canvases, token metadata, and adaptive temporal groups:
\begin{equation}
\label{eq:codec_interface}
\mathfrak{C}(\mathcal{V})=(\mathcal{X},\mathcal{U},\mathcal{G}),\qquad
\mathcal{X}=\{(X_s,\ell_s)\}_{s=0}^{S-1},\ \ell_s\in\{\mathrm{I},\mathrm{P}\},\qquad
\mathcal{U}=\{u=(\iota_u,f_u,\mathbf{p}^{\mathrm{can}}_u,\mathbf{p}^{\mathrm{src}}_u,\kappa_u)\}_{u=1}^{N}.
\end{equation}
Here \(\mathcal{X}\) contains \(S\) I/P canvases, \(\mathcal{U}\) contains \(N\) visual-token records, and \(\mathcal{G}=\{\mathcal{G}_k\}\) denotes the induced codec groups.
For token \(u\), \(\iota_u\) is the canvas index, \(f_u\) is the source-frame id, \(\mathbf{p}^{\mathrm{can}}_u\) is the packed canvas coordinate, \(\mathbf{p}^{\mathrm{src}}_u\) is the source-frame patch coordinate, and \(\kappa_u\) is the bit-cost-adaptive group id.
The packed coordinate supports compact canvas construction, while the source coordinate preserves the spatial origin of each token for spatiotemporal encoding.
The connector and language model do not consume these codec fields directly; codec-stream tokenization affects the model by selecting visual evidence and assigning token visibility groups.

\paragraph{Groups of Pictures (GOPs) Partition.}
Rather than assigning visual slots by elapsed time, codec-stream tokenization partitions video according to the temporal bit-cost profile of the compressed stream.
We divide the video into \(B\) bins of duration \(\Delta\) and aggregate the packet size of predicted frames within each bin:
\begin{equation}
\label{eq:pb_energy}
e_b=\sum_{q\in\mathcal{P}_{\mathrm{PB}}}
\operatorname{bytes}(q)\,
\mathbf{1}\{\tau(q)\in[b\Delta,(b+1)\Delta)\},
\qquad
\theta=\frac{\sum_{b=0}^{B-1}e_b}{\max(1,K_{\mathrm{tar}})} .
\end{equation}
Here \(\mathcal{P}_{\mathrm{PB}}\) is the set of P/B-frame packets, \(\operatorname{bytes}(q)\) is the packet size used as a proxy for prediction bit-cost, \(\tau(q)\) is the presentation timestamp of packet \(q\), and \(K_{\mathrm{tar}}\) is the target number of temporal groups.
Thus, \(e_b\) is a bin-level bit-cost rather than a per-frame score, and \(\theta\) is the average P/B bit-cost quota per adaptive GOP.
I-frame packets are excluded because they mainly reflect intra-frame spatial complexity, whereas P/B packets expose inter-frame prediction difficulty, motion, and residual change.

\begin{figure*}[t]
\centering
\includegraphics[width=0.9\textwidth]{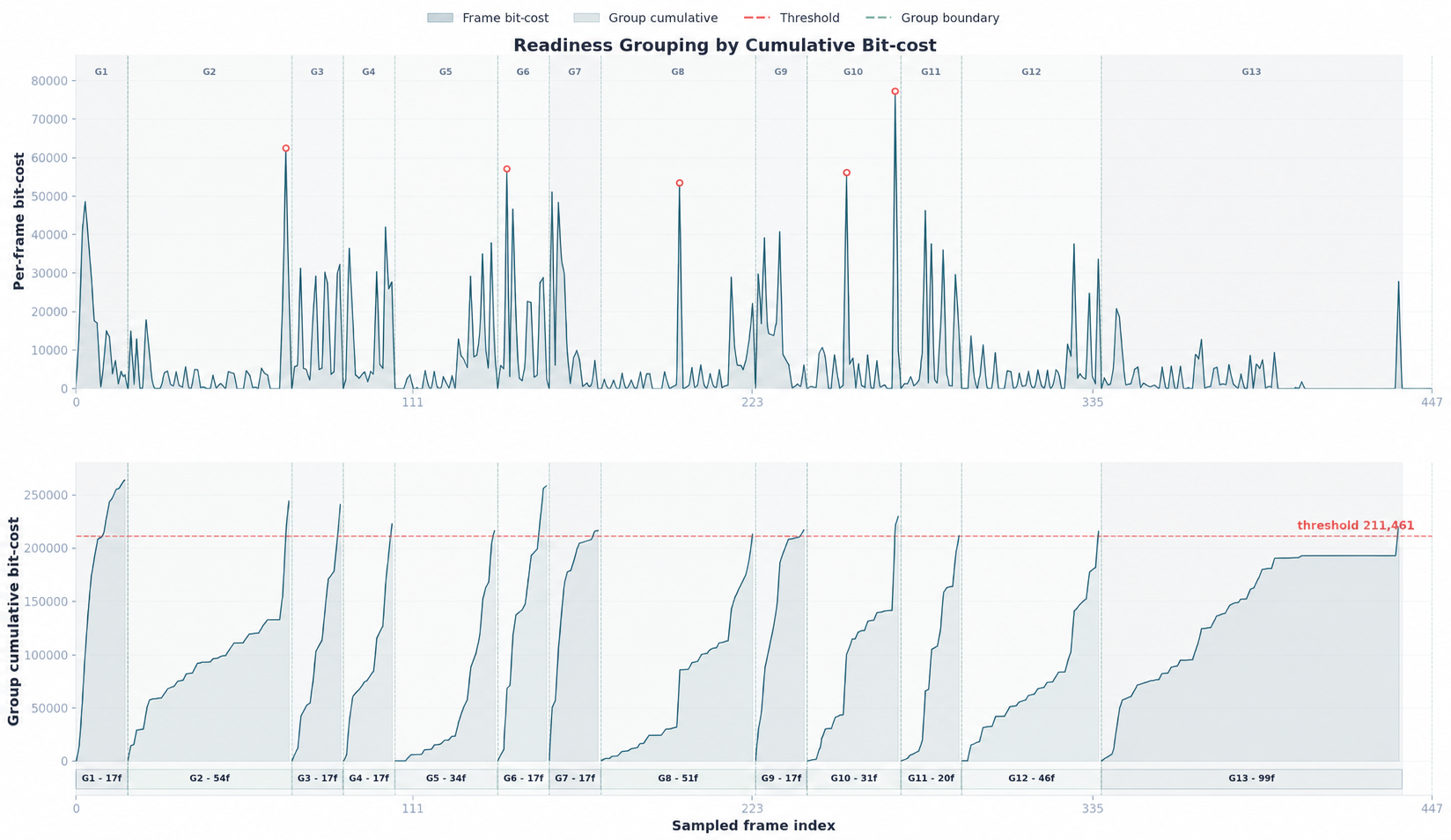}
\caption{\textbf{Codec-stream grouping by cumulative bit-cost.} For example, \(448\) sampled frames are divided into \(13\) codec-stream groups under a cumulative bit-cost threshold of \(211{,}461\). The top panel shows the frame-level bit-cost contribution in blue, where sharp peaks typically indicate rapid motion, viewpoint changes, or abrupt visual transitions. The bottom panel shows the cumulative bit-cost within each group in orange, which resets after every group boundary and approaches the red threshold before a new group is opened. Green dashed lines mark the resulting codec-stream group boundaries, and the bottom color bands indicate the number of frames covered by each group.}
\label{fig:stream_codec}
\end{figure*}

Starting from bin \(s_k\), the next boundary is triggered when the current segment either reaches the maximum span or accumulates sufficient bit-cost after the minimum span:
\begin{equation}
\label{eq:gop_trigger}
i_k=
\min\left\{
i\ge s_k:
(i-s_k+1\ge L_{\max})
\vee
\left[
i-s_k+1\ge L_{\min}
\wedge
\sum_{b=s_k}^{i}e_b\ge\theta
\right]
\right\},
\end{equation}
where \(L_{\min}=\lceil T_{\min}/\Delta\rceil\) and \(L_{\max}=\lceil T_{\max}/\Delta\rceil\) are the minimum and maximum group spans measured in bins.
The tentative boundary \(i_k\) is then refined by local valley search:
\begin{equation}
\label{eq:gop_valley}
c_k=\operatorname*{arg\,min}^{\mathrm{lex}}_{b\in\mathcal{N}_k(i_k)}(e_b,|b-i_k|),
\qquad
\mathcal{G}_k=[s_k,c_k],
\qquad
s_{k+1}=c_k+1 .
\end{equation}
The search window \(\mathcal{N}_k(i_k)\) is centered around \(i_k\) and constrained by the minimum span, maximum span, and video endpoints.
The lexicographic rule first selects the lowest-bit-cost valley and then chooses the closest bin to the trigger point.
High-change intervals therefore reach the quota quickly and form shorter groups, while predictable intervals span longer groups.
A token \(u\) receives \(\kappa_u=k\) if its source-frame time \(f_u/\mathrm{fps}\) falls inside \([s_k\Delta,(c_k+1)\Delta)\).
Figure~\ref{fig:stream_codec} visualizes this bit-cost-based adaptive grouping process.

\begin{figure*}[!t]
\centering
\includegraphics[width=\linewidth]{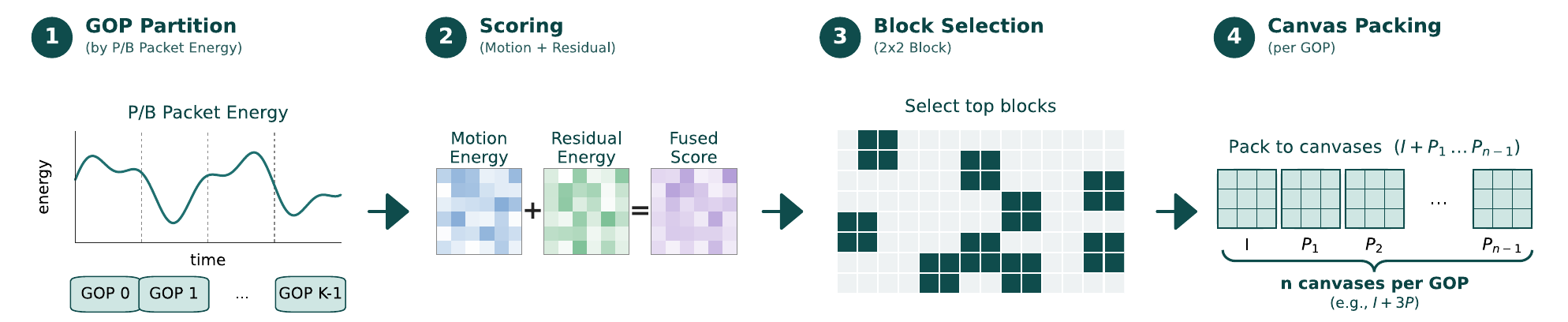}
\caption{\textbf{Codec-stream tokenization.}
P/B packet bit-cost partitions the video into adaptive GOPs; motion and residual signals jointly score spatial saliency; high-score \(2{\times}2\) patch blocks are selected and packed into compact I/P canvases.
Each GOP yields one anchor I-canvas and multiple P-canvases carrying motion-residual evidence, producing merge-aligned visual tokens whose density follows the bit-cost--residual profile of the stream rather than fixed frame slots.}
\label{fig:codec_stream_pipeline}
\end{figure*}

\paragraph{Scoring and Block Selection.} Within each bit-cost-adaptive group, motion-residual evidence determines which spatial regions are preserved. For a predicted frame \(t\), the codec exposes motion vectors \(\mathbf{d}_t\) and a luma residual \(r_t^Y\). As in the OV-Encoder, the motion field is densified to the pixel grid, the residual is interpreted around its zero point \(128\), and the two signals are normalized by robust percentile statistics. This gives a dense saliency map \(S_t(\mathbf{x})=\overline{M}_t(\mathbf{x})+\overline{R}_t(\mathbf{x})\), where \(\overline{R}_t\) denotes the normalized residual-response map, obtained by measuring the absolute luma-residual deviation from the zero point \(128\), scaling it by a robust frame-level percentile, and clipping it to \([0,1]\). Likewise, \(\overline{M}_t\) is the percentile-normalized motion-magnitude map derived from the densified codec motion vectors.


The difference from the original OV-Encoder patch mask is the selection granularity. 
Instead of selecting individual high-score patches, the codec patch-GOP path aggregates saliency into \(2{\times}2\) patch blocks. 
Let \(\mathcal{P}_{h,w}\) denote the \(p{\times}p\) region of patch \((h,w)\), with \(p=16\) in our implementation. 
The block score is \(A_{t,i,j}=\sum_{\alpha,\beta\in\{0,1\}}\sum_{\mathbf{x}\in\mathcal{P}_{2i+\alpha,\,2j+\beta}}S_t(\mathbf{x})\). 
Thus, a selected unit always contains the four neighboring patches \((2i,2j)\), \((2i,2j+1)\), \((2i+1,2j)\), and \((2i+1,2j+1)\). 
This block-level primitive is aligned with the encoder-side \(2{\times}2\) merge operation: every selected block contributes four spatially coherent patch tokens, avoiding the downstream merging of unrelated patches from different source regions or frames. We further augment the motion-residual score with a normalized patch-level bit-cost prior. Since bit-cost is naturally available at block granularity, it is fused during 2×2 block scoring rather than projected back to the pixel grid. The bit-cost term reflects local coding complexity and complements motion and residual energy for codec-aware spatial token allocation.

\paragraph{Canvas Packing.}
A global top-ranked selection over an entire codec group can over-concentrate tokens on a single high-response frame.
We therefore construct P-canvases through stratified temporal allocation.
Let \(\mathcal{Z}_k=\{z=(t,i,j,A_{t,i,j})\}\) be the candidate \(2{\times}2\) patch blocks inside group \(k\), where \(t\) is the source frame, \((i,j)\) is the block coordinate, and \(A_{t,i,j}\) is the block saliency score.
For each frame \(t\), we sort its candidate blocks by \(A_{t,i,j}\) in descending order and denote by \(\rho_t(z)\) the zero-based rank of candidate \(z\) within that frame.
We then attenuate repeated candidates from the same frame:
\begin{equation}
\label{eq:stratified_p_canvas}
\widetilde{A}_z=
\frac{A_{t,i,j}}{\sqrt{1+\lambda\rho_t(z)}},
\qquad
w_t=
\sum_{z\in\mathcal{C}_k(t)}\max(0,\widetilde{A}_z)
+
\alpha_{\mathrm{peak}}
\max_{z\in\mathcal{C}_k(t)}\widetilde{A}_z .
\end{equation}
Here \(\mathcal{C}_k(t)\subset\mathcal{Z}_k\) is the set of candidate blocks from frame \(t\) within group \(k\), \(\lambda\) controls the strength of same-frame attenuation, \(\alpha_{\mathrm{peak}}\) weights the strongest frame-level response, and \(w_t\) is the resulting frame-level allocation mass.
The attenuation prevents a single high-response frame from dominating the entire group, while the peak term preserves frames that contain a highly localized but important response.

To assign P-canvases across time, we sort candidate frames in group \(k\) as \(\{t_n\}_{n=0}^{M_k-1}\) and compute the cumulative allocation curve
\begin{equation}
\label{eq:temporal_allocation_curve}
F_k(\ell)=
\frac{\sum_{n=0}^{\ell}w_{t_n}}
{\sum_{n=0}^{M_k-1}w_{t_n}},
\qquad 0\le \ell < M_k .
\end{equation}
This curve maps the temporal order within group \(k\) to the fraction of accumulated saliency mass.
If group \(k\) is assigned \(m_k\) P-canvases, the \(r\)-th P-canvas, \(r\in\{0,\ldots,m_k-1\}\), draws high-scoring non-duplicate blocks from the frames whose cumulative allocation mass falls in \([r/m_k,(r+1)/m_k)\).
When the corresponding interval contains too few candidates, the selector expands to neighboring frames and finally falls back to the full group.
Thus, bit-cost dynamics determine where temporal resolution is needed, while motion-residual saliency and frame-level allocation weights determine how each group is covered by P-canvases.

\paragraph{Group-visible Attention.}
Codec-stream inputs, sampled-frame inputs, and static images share the same patch embedding and 3D positional encoding, forming a unified spatiotemporal token space. The OneVision-Encoder then uses a non-causal group-visible attention interface to define token visibility: sampled-frame and IPPP-style inputs use fixed four-slot groups, codec-stream inputs use the bit-cost-adaptive GOP id \(\kappa_u\) so tokens from the same variable-length GOP remain group-visible across P-canvases, and static images reduce to a single-temporal group. Consequently, all input forms share the same encoder parameters, with only evidence allocation and group assignment varying across inputs.       
\section{Training Data}
\label{sec:training_data}

\begin{figure*}[t]
  \centering
  \begin{subfigure}[t]{0.48\linewidth}
    \centering
    \includegraphics[width=\linewidth]{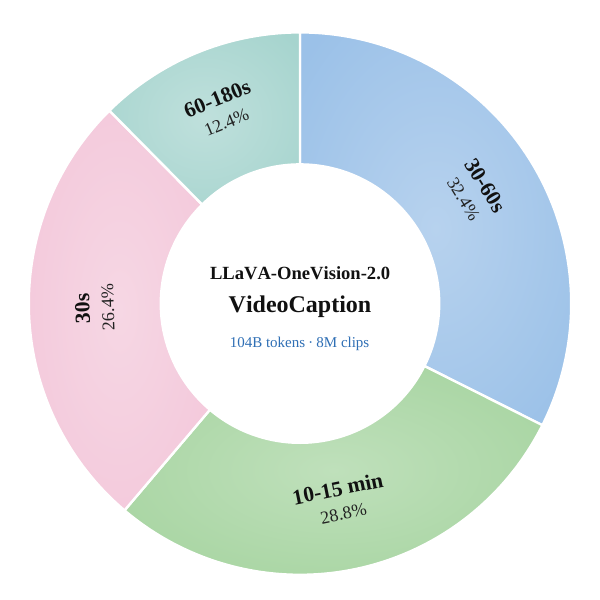}
    \caption{Video-Caption Data Mixture}
    \label{fig:lov2-videocaption}
  \end{subfigure}
  \hfill
  \begin{subfigure}[t]{0.48\linewidth}
    \centering
    \includegraphics[width=\linewidth]{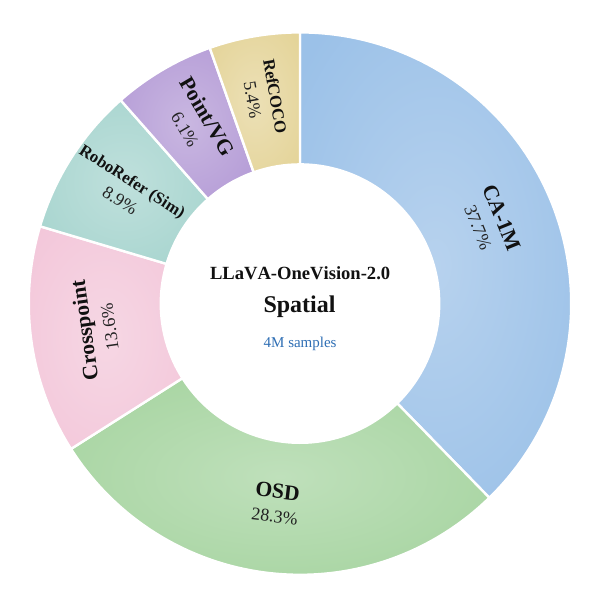}
    \caption{Spatial Data Mixture}
    \label{fig:lov2-spatial}
  \end{subfigure}
  \caption{\textbf{LLaVA-OneVision-2 data mixtures.} (a) Token-volume proportions of the video-caption corpus and (b) the spatial-reasoning corpus used during training. The video-caption mixture contains 104.1B tokens from 7.96M clips spanning four duration buckets, while the spatial mixture aggregates 4M samples drawn from six datasets covering 3D scenes, spatial reasoning, pointing, and referring expressions.}
  \label{fig:lov2-mixtures}
\end{figure*}

The LLaVA-OneVision-2 recipe consumes data from three buckets, each contributing a distinct slice of the model's eventual capability surface. 
\subsection{Inherited Image--Text Foundation}
\label{subsec:inherited_data}

\paragraph{Image--text foundation.}
We initialise from the image-pretrained checkpoint of LLaVA-OneVision-1.5 and reuse the LLaVA-OneVision-1.5 mid-training and instruction corpora as-is. The mid-training corpus (LLaVA-OneVision-1.5-Mid-Training-85M) is concept-balanced over $\sim$85M image--text pairs (20M~ZH + 65M~EN); the instruction corpus (LLaVA-OneVision-1.5-Instruct-Data, $\sim$22M samples) covers OCR, GUI, document, grounding, counting, and chart/diagram tasks. We additionally include FineVision ($\sim$24M instruction samples) for broader image-instruction coverage. Detailed per-source statistics for the mid-training and instruction corpora are reported in the LLaVA-OneVision-1.5 release; we do not re-derive or modify them here.

\paragraph{Inherited video instruction.}
For video instruction tuning, we utilize relevant subsets from four publicly available corpora: LLaVA-Video-178K~\citep{zhang2024llavavideo} (1.6M samples covering captioning, open-ended QA, and multiple-choice QA), VideoChat-Flash-Training-Data~\citep{li2025videochatflash}, Molmo2~\citep{molmo2,clark2026molmo2}, and TimeLens. Only the data pertinent to our methodology is selected from each corpus. These corpora are general video-instruction data rather than long-form sources, and we deliberately do not synthesize any additional long-video instruction data: all long-context capability is acquired from the length-stratified caption corpus, while the inherited corpora supply instruction-following diversity.

\subsection{Length-Stratified Video Caption Corpus}
\label{subsec:video_caption_corpus}

\paragraph{Length stratification as a design choice.}
A central component of our training data is a length-stratified video caption corpus spanning 30 seconds to 15 minutes, totalling approximately 8M captioned clips. We deliberately stratify by length because uniform-length captioning recipes (typically dominated by short clips) over-represent semantic perception relative to temporal continuity: the model learns to ``describe a scene'' but not to ``maintain state across ten minutes.'' The four buckets (30s, 30--60s, 60--180s, 10--15min) are sized so that each successive stage of the training recipe (\S\ref{sec:training_recipe}) can extend its frame budget by a factor of 2--4$\times$ without out-running its caption supervision.
We compute image tokens at $392{\times}392$ input with ViT patch size 14 and a $2{\times}2$ vision merge, yielding 196 visual tokens per frame; caption tokens are measured with the Qwen3 tokenizer over a 1{,}500-sample average per bucket and then scaled by row count.

\paragraph{Codec-aware re-encoding at Stage 4.}
The 10--15-minute bucket is consumed twice in Stage 4 (\S\ref{subsec:stage4}): once at 384 frames under the variable-length-GOP, bit-cost-scored codec configuration of \S\ref{sec:codec_pipeline}, and once at 768 frames under the same configuration to densify the temporal axis at the upper end of the recipe's frame-budget schedule. No new captions are produced for the densified pass; the same per-clip caption is re-aligned against a denser visible-patch index.
\section{Training Recipe}
\label{sec:training_recipe}

The LLaVA-OneVision-2 recipe runs in four progressive stages (\S\ref{subsec:stage1}--\S\ref{subsec:stage4}).

\subsection{Stage 1 -- Bootstrap from LLaVA-OneVision-1.5 + 30s Video Captions}
\label{subsec:stage1}

We initialize from the image-pretrained LLaVA-OneVision-1.5~\citep{an2025llava} 8B checkpoint and bootstrap it into a video-aware model by mixing in short video-caption data. 
The training corpus consists of (i) LLaVA-OneVision-1.5-Mid-Training-85M~\citep{an2025llava}, a concept-balanced image--text dataset, and (ii) our newly released \newrelease{30s-Video-Caption-4.2M} corpus, where each sample is paired with a caption over a 0--30s video span, sampled at 1 fps with up to 30 frames. 
All video inputs in this stage are constructed with standard frame sampling; codec-stream tokenization is not used.

\subsection{Stage 2 -- Instruction Tuning + 30--60s Video Captions}
\label{subsec:stage2}

Stage 2 introduces large-scale multimodal instruction data and extends video understanding to medium-length clips. 
The training mixture consists of (i) LLaVA-OneVision-1.5-Instruct-Data (\(\sim\)22M samples)~\citep{an2025llava}, (ii) HuggingFaceM4/FineVision (\(\sim\)24M samples), (iii) our newly released \newrelease{30s--60s-Video-Caption-2.7M} corpus, where each example is paired with a caption over a 30--60s video span sampled at 1 fps with up to 60 frames, and (iv) our newly released \newrelease{60s--180s-Video-Caption-700K} corpus, where each example is paired with a caption over a 60--180s video span using up to 90 frames. 
The maximum frame budget is increased from 30 to 90 in this stage, marking the first expansion step in the cross-stage schedule described in \S\ref{subsec:training_dynamics}. 
All video inputs in this stage continue to use standard frame sampling without codec-stream tokenization.

\subsection{Stage 3 -- Long Video Understanding}
\label{subsec:stage3}

Stage 3 extends training to long-form video understanding by combining long-duration caption data with established video instruction corpora. 
The training mixture includes LLaVA-OneVision-1.5-Instruct-Data, HuggingFaceM4/FineVision, lmms-lab/LLaVA-Video-178K~\citep{zhang2024llavavideo}, OpenGVLab/VideoChat-Flash-Training-Data~\citep{li2025videochatflash}, and our newly released \newrelease{10min--15min-Video-Caption-350K} corpus, which pairs captions with 10--15 minute video spans. 
The maximum frame budget is increased from 90 to 384 in this stage. 
As in the previous stages, all video inputs are constructed with standard frame sampling.

\subsection{Stage 4 -- Extended Video and Codec-Stream Training}
\label{subsec:stage4}

Stage 4 further extends long-video training and introduces dedicated spatial supervision. Up to Stage 3, the recipe relies predominantly on caption-style supervision, which provides only indirect learning signals for fine-grained spatial structure. In the final stage, codec-stream tokenization is introduced for the \newrelease{10min--15min-Video-Caption-350K} corpus. Specifically, we revisit this long-video caption corpus with the variable-length-GOP, bit-cost-scored codec pipeline described in \S\ref{sec:codec_pipeline}. The corpus is used both at up to 384 frames and in a densified variant with up to 768 frames under the same codec-stream configuration, making it the largest visual-budget caption component in the training recipe. All remaining Stage-4 data, including multimodal instruction-tuning data, spatial QA data, Molmo2-VideoTrack, and Molmo2-VideoPoint, continues to use the standard input format of the corresponding data source and does not apply codec-stream tokenization. Therefore, codec-stream is not a global Stage-4 preprocessing rule; it is a targeted input representation for the long-video caption component.

Stage 4 also incorporates two spatially oriented data sources. Our in-house \newrelease{LLaVA-OneVision-2-Spatial-4M} corpus provides 4M structured QA pairs covering size, direction, count, distance, and appearance order over annotated indoor scans, embodied-simulator trajectories, and pseudo-annotated web video frames. We further adopt Molmo2-VideoTrack and Molmo2-VideoPoint~\citep{molmo2} for point-based tracking and spatio-temporal pointing. Together, these datasets extend the supervision signal from static spatial relations to temporally grounded spatial understanding.

\section{Implementation Details}
\label{subsec:training_dynamics}

Three cross-stage design choices are shared across all four stages and cannot be ablated by reading any single stage's data list. Three cross-stage design choices govern the LLaVA-OneVision-2 training recipe beyond the data mixture of any individual stage:

\paragraph{Mixed-batch Composition.}
Each training step interleaves three input forms from the unified OneVision-Encoder interface: codec-stream videos, uniformly sampled videos, and static image inputs. 
In practice, the mixture is approximately \(50\%\) codec-patchified video, \(37.5\%\) uniform chunk-wise video, and \(12.5\%\) image inputs. 
This composition exposes the model to both stream-aware event allocation and frame-faithful visual evidence, while preserving image-understanding capability. 
Because all input forms share the same patch embedding, 3D RoPE, and visible-token interface, the mixture requires no modality-specific routing or additional input-form parameters.

\paragraph{Frame-budget Schedule.}
The visible per-clip frame budget increases progressively across stages: 30 frames in Stage 1, 60/90 frames in Stage 2, 384 frames in Stage 3, and 384/768 frames in Stage 4. 
This schedule matches the temporal span of the supervision: short-video captions support early-stage perception, while long-video captions and codec re-encoding support dense long-form observation in later stages. 
Since training does not exceed 768 visible frames, denser or hour-scale inputs remain an extrapolation regime for future streaming and ultra-long-context codec modeling.

\paragraph{Cross-stage Codec Scheduling.}
Stages 1--3 use standard frame-sampling inputs and do not apply codec-stream tokenization. Codec-stream training is introduced in Stage 4, where we switch the long-video data to the variable-length-GOP, bit-cost-scored codec pipeline described in \S\ref{sec:codec_pipeline}. The 10--15 minute video-caption corpus is re-encoded under this Stage-4 configuration at 384 and 768 frames. Because codec-stream is enabled only in the final stage, we evaluate codec-stream versus uniform frame sampling under the Stage-4 setting, where the two input strategies share the same model, supervision data, training schedule, and evaluation protocol.
\section{JumpScore}
\label{sec:jumpscore}

\begin{figure*}[!t]
    \centering
    \includegraphics[width=\linewidth]{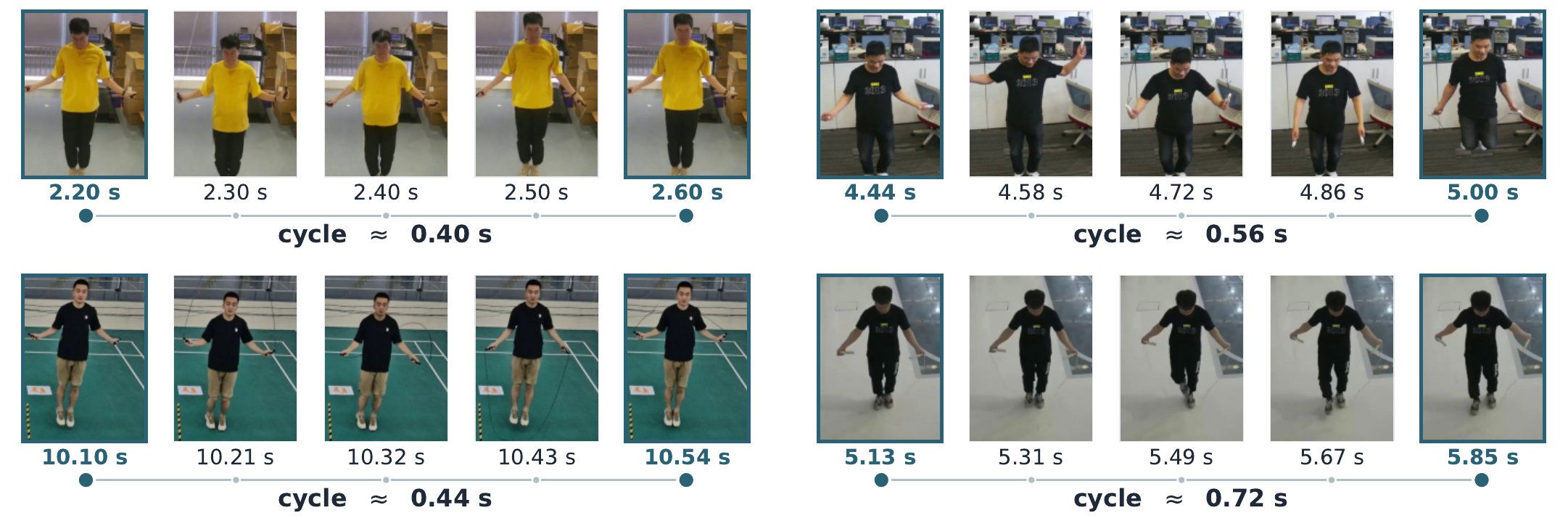}
    \caption{\textbf{Representative cycles from the JumpScore benchmark.} Four clips, each decomposed into five frames spanning one jump-rope cycle. The first and last frame of every panel are ground-truth cycle starts (rope behind legs). The four panels span warehouse, office, sports-court, and tiled-corridor captures.}
    \label{fig:jumpscore_cycle}
\end{figure*}

Standard temporal-grounding benchmarks such as Charades-STA~\citep{gao2017tall}, ActivityNet~\citep{caba2015activitynet}, and QVHighlights~\citep{lei2021detecting} target one-shot event localization, where adjacent visual evidence is already distinguishable. \textbf{JumpScore} probes the opposite regime: fine-grained grounding in high-frequency, densely repeating motion where adjacent cycles are visually near-identical and the discriminative signal lives at cycle boundaries. The benchmark consists of 189 in-the-wild jump-rope videos with complete decimal-second annotations of every cycle start, defined by the moment the rope passes behind the legs. Figure~\ref{fig:jumpscore_cycle} shows four representative clips spanning the range of scenes, camera angles, and cycle periods in the benchmark. The dataset is publicly released on Hugging Face\footnote{\url{https://huggingface.co/datasets/lmms-lab-encoder/JumpScore}} and is integrated into the \texttt{lmms-eval-ov2} evaluator used elsewhere in this paper.

\paragraph{Dataset Construction.}
The 189 source videos span multiple indoor scenes, camera angles, and capture devices, with clip durations concentrated in the 30--90\,s range so the typical clip contains tens of cycles at sub-second spacing. Capture resolution is at least 1280$\times$720 on every clip and 1920$\times$1080 or higher on more than 84\% of them, keeping the rope-and-leg cue visually resolvable. Cycle-start timestamps are annotated at decimal-second precision and verified against the video's source frame rate so that the same boundary frame is recovered under deterministic decoding. Targeting a single, visually well-defined motion primitive isolates cycle-boundary localization from confounders such as category recognition, scene parsing, and text-and-dialogue cues, which is what makes sub-second annotation tractable across the full 189-clip set.

\paragraph{Task and Metric.}
Each video is paired with a fixed natural-language prompt: ``List the start timestamps in s of each jump rope the main character does in the video. The start is defined as the moment the rope is behind the legs.'' The expected output is a list of decimal-precision timestamps in seconds; no auxiliary modality is provided, and the same prompt is used for every video so that performance differences reflect grounding capability rather than prompt sensitivity. Predictions are matched greedily to ground-truth timestamps within a temporal tolerance band of $\delta$ seconds: a predicted timestamp counts as a true positive if it falls within $\delta$ of an unmatched ground-truth timestamp, remaining predictions count as false positives, and unmatched ground-truth timestamps count as false negatives. The headline metric is mean Average Precision (mAP) averaged across three sub-second tolerance levels, $\delta\in\{0.1,0.2,0.3\}$\,s. Against a median cycle period of roughly 0.4\,s, AP@0.1 is the strict regime, in which a correct match must land within a half-cycle window of its referent; AP@0.2 and AP@0.3 are looser and admit some drift toward neighboring cycles. We report the mean across all three so the metric grades both precise cycle attribution and coarser localization, with AP@0.1 as the strictest of the three. The mAP aggregation jointly rewards completeness across the full cycle sequence and precision in placing each prediction near the correct cycle, and penalizes degenerate strategies such as enumerating evenly spaced timestamps to cover the clip.
\section{Evaluation}
\label{sec:evaluation}

This section reports end-to-end results for LLaVA-OneVision-2 against 8B-class open MLLMs across three benchmark domains targeted by our recipe: \emph{video understanding}, \emph{spatial reasoning}, and \emph{image and document understanding} (\S\ref{subsec:main_results}). 
We further evaluate referring video object segmentation as a robotics-relevant test of temporally coherent spatial grounding and object-identity consistency. 
The codec-versus-uniform-sampling analysis underlying the video gains is reported separately in \S\ref{sec:codec_analysis}.

\mainvideobenchmarktable

\subsection{Main Multimodal Benchmark Results}
\label{subsec:main_results}

We evaluate \textbf{LLaVA-OV-2-8B} on a comprehensive suite of public benchmarks spanning \emph{video understanding}, \emph{spatial reasoning}, and \emph{image and document understanding}. 
We compare against four 8B-class baselines: Qwen3-VL-8B, Keye-VL-1.5-8B, InternVL-3.5-8B, and LLaVA-OV-1.5-8B; PLM-8B is included where results are available. 
Unless otherwise specified, all numbers are obtained with LMMs-Eval~\citep{lmms_eval} using default prompts, identical decoding strategies, and matched token budgets across compared models.

\mainspatialimagebenchmarktable

\paragraph{Video Understanding.}
Table~\ref{tab:video_main} reports results on 18 video benchmarks covering short-clip QA (MV-Bench, NextQA, TempCompass), full-length VideoMME (with and without subtitles, plus VideoMME-v2~\citep{fu2026videommev2}), long-video understanding (MLVU-dev, LVBench, LongVideoBench), expert and omni-modal reasoning (MMVU-val, VideoEval-Pro, MMOU), temporal grounding (Charades-STA, ActivityNet, QVHighlights), visual--spatial intelligence in video (VSI-Bench~\citep{yang2024think}, ReVSI~\citep{zhang2026revsi}), and \textbf{JumpScore}, for which we report JumpScore mAP. 
JumpScore targets cycle-level temporal localization in high-frequency repetitive motion, where models must distinguish the correct action instance among many visually similar motion cycles. 
LLaVA-OneVision-2-8B achieves the highest average score across this 18-task suite, improving over Qwen3-VL-8B by \(+4.3\) points (62.5 vs.\ 58.2). 
The largest gains appear on the axes targeted by our recipe: temporal grounding (Charades-STA: \(+5.2\), ActivityNet: \(+7.0\), QVHighlights: \(+7.0\)), visual--spatial video reasoning (VSI-Bench: \(+11.8\), ReVSI: \(+8.7\)), and JumpScore (\(74.9\) vs.\ \(30.1\) JumpScore mAP, \(+44.8\) over Qwen3-VL-8B). 
These results indicate that codec-stream tokenization and shared 3D RoPE improve event-level video evidence allocation under irregular spatiotemporal token layouts.

\finding{}{\faBookmark~~LLaVA-OneVision-2-8B leads the 8B class on the 18-task video suite, with the largest gains on temporal grounding, visual--spatial video reasoning, and JumpScore temporal localization.}

\paragraph{Spatial Reasoning.}
We next evaluate spatial reasoning, the axis most directly targeted by the Stage-4 spatial recipe. 
Table~\ref{tab:spatial_main} reports 11 spatial benchmarks spanning relation comprehension, embodied scene understanding, 2D/3D spatial reasoning, trajectory reasoning, cross-view correspondence, spatial aptitude, multi-image spatial intelligence, and general perception. 
LLaVA-OneVision-2-8B achieves the best spatial average, improving over Qwen3-VL-8B by \(+5.3\) points on this 11-task subset (63.5 vs.\ 58.2). 
Its largest gains occur where existing 8B baselines are weakest: CrossPoint improves by \(+35.0\) points over Qwen3-VL-8B (61.9 vs.\ 26.9), and TraceSpatial-3D reaches nearly \(4\times\) the next-best 8B score (31.0 vs.\ 8.0). 
The model further leads or matches the 8B class on MetaVQA, CV-Bench 2D/3D, EmbSpatial, and SAT, while remaining competitive on CRPE, ERQA, MMSI-Bench, and BLINK. 
We attribute these gains to the Stage-4 spatial corpus and point-based tracking supervision, which explicitly train structured spatial outputs beyond caption-style perception.

\finding{}{\faBookmark~~Stage-4 spatial supervision yields the largest gains where 8B baselines are weakest: \(+35.0\) on CrossPoint and nearly \(4\times\) the next-best 8B score on TraceSpatial-3D.}

\paragraph{Image and Document Understanding.}
We also evaluate whether long-video and spatial training preserve image and document capability. 
Table~\ref{tab:spatial_main} summarises results on 11 image and document benchmarks. 
LLaVA-OneVision-2-8B remains competitive on DocVQA (95.2), ChartQA (85.9), CountBench (89.0), Pixmo-Count (64.0), and RealWorldQA (69.7), and leads the 8B class on V$^{*}$-Bench (85.9). 
On text-dense or diagram-heavy tasks, it trails the strongest specialized baselines: Keye-VL-1.5 leads MMStar and OCRBench, Qwen3-VL-8B leads InfoVQA, and PLM-8B leads AI2D. 
These gaps indicate that the spatially grounded long-video recipe preserves strong image-level capability, but does not make the model OCR- or document-specialized.

\finding{}{\faBookmark~~LLaVA-OneVision-2 preserves strong image and document capability, leading V$^{*}$-Bench while remaining below OCR- and document-specialized baselines on text-dense tasks.}         
\section{Ablation of Codec-Stream Tokenization}
\label{sec:codec_analysis}

\begin{figure*}[!t]
    \centering
    \includegraphics[width=\linewidth]{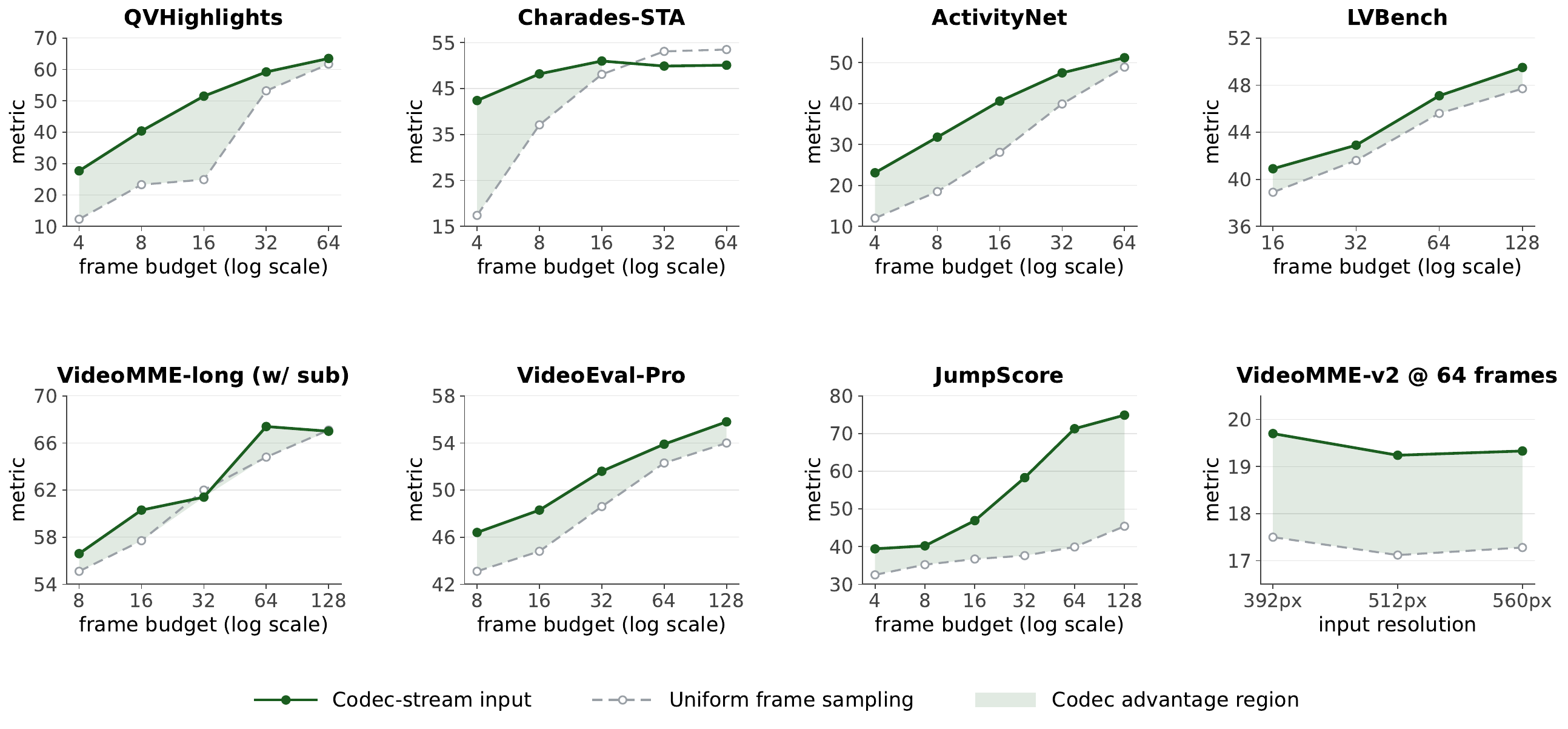}
    \caption{\textbf{Codec-stream inputs versus frame sampling.}
    Under matched visual settings, codec-stream tokenization improves event-level temporal grounding by following high-bit-cost intervals and high-residual regions rather than uniformly sampled frame slots.
    We evaluate this effect on temporal grounding benchmarks, long-form video QA, and JumpScore, our fine-grained temporal-localization benchmark for high-frequency repeated motion.}
    \label{fig:codec_vs_frame}
\end{figure*}

Codec-stream tokenization is designed to make video observation follow the predictive structure of the compressed stream. 
To isolate this effect, we compare codec-stream inputs with frame-sampling inputs while keeping the backbone, language model, decoder, prompts, and evaluation protocol fixed. 
The results show that codec-stream tokenization is most effective when the answer depends on event-level evidence: Codec-stream yields a 17.3-point average gain on JumpScore and a 9.7-point average gain across three temporal-grounding benchmarks over uniform frame sampling. 
On long-form video QA, codec-stream inputs preserve parity or small gains over frame sampling, indicating that stream-aware evidence allocation does not trade off long-video semantic understanding for compression.

\paragraph{Temporal Grounding.}
Temporal grounding exposes the core advantage of codec-stream tokenization. 
Uniform frame sampling allocates observation capacity by elapsed time, so short event intervals can be missed when they fall between sampled frames. 
Codec-stream tokenization instead uses bit-cost dynamics to identify temporally informative segments and motion-residual cues to retain spatial evidence around perceptual transitions. 
As shown in Figure~\ref{fig:codec_vs_frame}, this produces the largest gains at low input scales, where uniform sampling has the highest probability of missing event boundaries. 
The gap narrows as the input scale increases, since denser frame sampling gradually recovers more event evidence. Averaged over all temporal-grounding settings, codec-stream improves the score from 35.5 to 45.2, yielding a 9.7-point average absolute gain over uniform frame sampling.

\finding{}{\faBookmark~~Codec-stream tokenization improves temporal grounding by biasing visual evidence toward high-bit-cost transitions and high-residual regions, rather than distributing tokens uniformly over time.}

\paragraph{Fine-grained Grounding in Densely Repeated Motion.}
JumpScore targets a complementary failure mode of existing long-video benchmarks: high-frequency, densely repeated motion, where the challenge is not recognizing the event category but localizing the correct action instance among many visually similar cycles. 
In this regime, periodic frames can appear redundant under visual similarity, even though their cycle boundaries carry the decisive temporal signal. 
Codec-stream tokenization is well matched to this setting because bit-cost spikes and residual responses concentrate around cycle transitions. 
On JumpScore, LLaVA-OneVision-2-8B reaches \(74.9\) JumpScore mAP, outperforming Qwen3-VL-8B (\(30.1\)) by \(+44.8\) points; under matched visual budgets on the same benchmark, codec-stream inputs improve temporal grounding over frame sampling by \(+17.3\) points.

\begin{figure*}[!t]
    \centering
    \includegraphics[width=\linewidth]{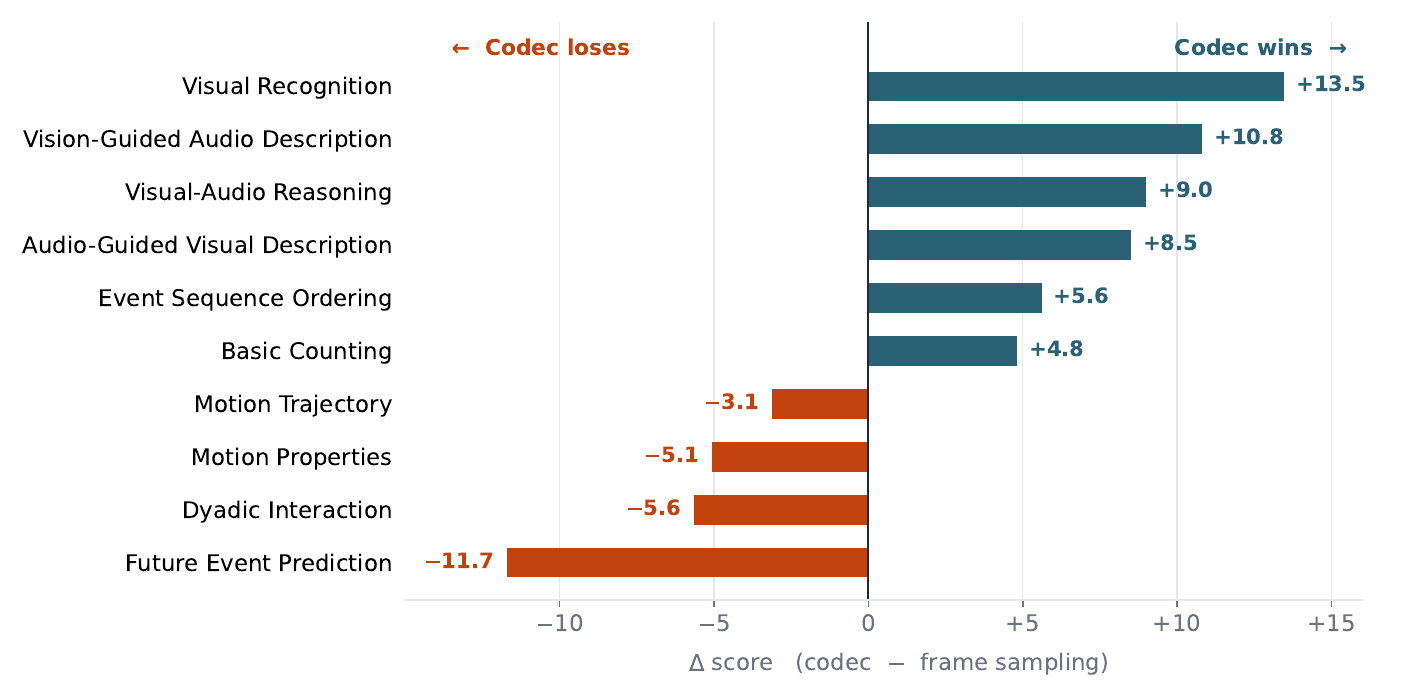}
    \caption{\textbf{Per-skill ablation on VideoMME-v2 at matched visual settings.}
    Teal bars indicate capabilities where codec-stream inputs outperform frame sampling; coral bars indicate capabilities where frame sampling is stronger.}
    \label{fig:codec_vs_frame_videommev2}
\end{figure*}

\finding{}{\faBookmark~~On densely repeated motion, codec-stream tokenization localizes perceptual transitions between visually similar cycles, the regime where uniform frame sampling and similarity-based deduplication are least reliable.}

\paragraph{Long-form Video Question Answering.}
Long-form video QA is a more forgiving setting for frame sampling, since many answers can be recovered from sparse semantic snapshots. 
The relevant question is therefore whether codec-stream tokenization introduces a regression by prioritizing motion-residual evidence. 
Figure~\ref{fig:codec_vs_frame} shows that it does not: codec-stream inputs maintain parity or small gains on LVBench, VideoMME-Long with subtitles, and VideoEval-Pro. 
This suggests that codec-stream tokenization preserves broad semantic coverage while reallocating additional capacity toward event-bearing moments. 
At higher input scales, the two strategies converge, indicating that abundant frame evidence can partially compensate for less adaptive allocation.

\finding{}{\faBookmark~~On long video QA, codec-stream tokenization preserves the semantic robustness of frame sampling while allocating additional evidence to stream-indicated event moments.}

Figure~\ref{fig:codec_vs_frame_videommev2} further clarifies the competence boundary of codec-stream tokenization. 
Codec-stream inputs gain on tasks whose decisive evidence is a salient spatial snapshot or a discrete state transition, including visual recognition, event ordering, counting, and several audio-visual reasoning skills. 
Frame sampling remains stronger when the task requires dense trajectory continuity, such as future event prediction, dyadic interaction, motion properties, and motion trajectory estimation. 
This pattern supports the coverage--resolution trade-off identified in our analysis: codec-stream inputs expand event-level coverage, while frame sampling preserves frame-level evidential resolution for detail-sensitive motion reasoning.

\longvideocodectable

\paragraph{Fixed Versus Stream-adaptive Codec Allocation.}
To isolate the contribution of stream adaptivity inside the codec pipeline, we compare \textsc{Fix}, which follows a fixed GOP/slot schedule, with \textsc{Stream}, which derives temporal groups and key-frame anchors from the continuous bit-cost stream and packs motion-residual evidence into compact visual canvases. 
Across LVBench, VideoMME-Long, MLVU-dev, and VideoEval-Pro, \textsc{Stream} matches or improves over \textsc{Fix} in most settings, with especially consistent gains on MLVU-dev and VideoEval-Pro. 
The result supports our central design claim: the advantage of codec-stream tokenization comes not merely from compression, but from making visual observation follow bit-cost dynamics and motion-residual structure rather than a preset GOP layout.    
\section{Related Work}
\label{sec:related_work}

LLaVA-OneVision-2 sits at the intersection of five lines of work: open video MLLMs (\S\ref{subsec:rw_video_mllm}), patch- and token-level efficiency for video transformers (\S\ref{subsec:rw_efficient_tok}), temporal grounding with video LLMs (\S\ref{subsec:rw_temporal_grounding}), spatial cognition in multimodal LLMs (\S\ref{subsec:rw_spatial}), and referring video object segmentation (\S\ref{subsec:rw_rvos}). We review each in turn and position codec-stream tokenization within them.

\subsection{Open Video MLLMs}
\label{subsec:rw_video_mllm}

The 8B-class open multimodal model has converged into a recognisable design template: a vision transformer backbone (typically SigLIP-style or a derivative), a connector projecting visual patches into the language-model embedding space, and an instruction-tuned language model (Qwen, LLaMA, InternLM). Recent representative releases at this scale include Qwen2-VL / Qwen2.5-VL / Qwen3-VL~\citep{wang2024qwen2,bai2025qwen25vl,Qwen3-VL}, InternVL-3 / InternVL-3.5~\citep{internvl3,wang2025internvl35}, Keye-VL-1.5~\citep{Keye-VL-1.5}, NVILA~\citep{liu2024nvila}, VideoLLaMA 3~\citep{zhang2025videollama3}, MiniCPM-V~\citep{yao2024minicpmv}, Aria~\citep{aria}, Eagle 2~\citep{chen2025eagle2}, Apollo~\citep{zohar2024apollo}, Oryx-MLLM~\citep{liu2024oryx}, Tarsier 2~\citep{xu2025tarsier2}, LongVU~\citep{longvu}, LongVILA~\citep{chen2024longvila}, InternVideo2.5~\citep{wang2025internvideo25}, Penguin-VL~\citep{zhang2026penguin}, and the LLaVA-OneVision-1.5 release~\citep{llava-onevision,an2025llava} that LLaVA-OneVision-2 inherits from directly. Most of these models share the same default observation strategy on video input: a clip is reduced to 8--32 uniformly-spaced frames, every patch in each sampled frame is encoded, and the rest of the timeline is discarded. The Cambrian-1~\citep{cambrian} and Cambrian-S~\citep{cambrian-s} releases additionally probe the failure modes of this default, particularly long-horizon continual reasoning, without altering the underlying frame-sampling prior. VideoChat~\citep{li2023videochat,Maaz2023VideoChatGPT}, ShareGPT4Video~\citep{chen2024sharegpt4video}, and the LLaVA-Video instruction-data line~\citep{zhang2024llavavideo,Maaz2023VideoChatGPT} are likewise built on top of uniform sampling. Our contribution is orthogonal to this template: codec-stream tokenization is a substitution at the input side of the visual encoder, leaving the connector, language model, and instruction-tuning recipe unchanged. Any of the systems above can in principle adopt it without touching their downstream stack.

\subsection{Efficient Video Tokenization}
\label{subsec:rw_efficient_tok}

A large body of prior work has tackled the same redundancy problem we identify, that most patches across most frames are not informative, but from a different direction. Three families of approaches deserve comparison.

\paragraph{Token dropout and merging.}
Token dropout~\citep{rao2021dynamicvit,yin2022adavit,meng2022adavit} removes a learned subset of tokens at intermediate transformer layers; token merging~\citep{bolya2022tome,bolya2023tomesd} combines similar tokens via attention-key similarity. A second family of recent work pushes this idea into the video-MLLM stack itself: hierarchical clip-to-video compression in VideoChat-Flash~\citep{li2025videochatflash}, one-token-per-frame summarisation in LLaVA-Mini~\citep{zhang2025llavamini}, Mamba-based temporal pooling in STORM~\citep{jiang2025storm}, dual-stream slow-fast token routing~\citep{xu2025slowfast}, training-free redundancy pruning in ReTaKe~\citep{wang2024retake}, and reconstructive or KV-sparsification compression in the Video-XL family~\citep{shu2025video,liu2025videoxlpro,shu2025videoxl2}. Concurrent 2026 entries continue this trajectory: visual KV-cache memory mechanisms (FlexMem~\citep{chen2026flexmem}), small-VLM-as-local-compressor pipelines (Tempo~\citep{fei2026tempo}), extreme one-token-per-frame compression~\citep{zhang2026xcomp}, density-adaptive samplers~\citep{chen2026compactvideo}, RL-learned per-token retention (SCORE~\citep{wang2026score}), semantic-guided evolutionary compression (EvoComp~\citep{song2026evocomp}), and hybrid KV-cache compression~\citep{zeng2026hybridkv}. All of these families typically operate at the model side (inside the transformer stack or the LLM-side KV cache) rather than at the input side, and require either learned routing, a similarity-based pooling step, or a task-conditional sparsification policy that adds inference-time computation. Codec-stream tokenization differs in two ways: it operates entirely at the input side, so any downstream model can be a plain transformer with no routing logic; and it derives the saliency signal from the codec bitstream rather than from learned model activations or post-hoc similarity, so the selection is computed once at preprocessing time and is agnostic to the downstream model.

\paragraph{Compressed-domain video tokenization.} A closer line of work directly exploits canonical codec representations as token sequences. Unlike JPEG-LM~\citep{han2024jpeg}, which models compressed file bytes for visual generation, our codec-stream tokenization uses the codec bitstream as an input-side saliency signal for video understanding and patch selection under a fixed token budget. Video-LaVIT~\citep{jin2024video} decomposes videos into keyframes and motion vectors for unified generative pretraining. Run-Length Tokenization (RLT)~\citep{choudhury2025rlt} exploits temporal redundancy at the token level. EMA~\citep{zhao2025efficient} encodes GOP structures with motion-aware mechanisms for efficient video MLLM understanding. AuroraLong~\citep{xu2025auroralong} brings RNN-style token compression back for efficient long-form video understanding. Concurrent 2026 work has further explored this direction: ReMoRa~\citep{yashima2026remora} operates on sparse RGB I-frames augmented with refined motion-vector representations through a hierarchical motion state-space module, and a recent survey~\citep{jin2026compression} argues for the joint standardisation of traditional visual coding (H.264/H.265) and visual token technology. Each of these methods extracts a different signal from the compressed bitstream and demonstrates a different efficiency--accuracy trade-off; what is missing in all of them is a unified \emph{operational} pipeline that runs at training and inference under the same configuration, makes the signal-extraction choices explicit and tunable (variable-length GOP, bit-cost vs.\ MV+residual scoring, stratified block selection), and Pareto-dominates uniform sampling across temporal grounding, R-VOS, and 3D-RoPE token-efficiency under matched token budgets. \S\ref{sec:codec_pipeline} provides exactly that pipeline; \S\ref{sec:codec_analysis} provides the controlled empirical comparison across temporal grounding, counting, long-form QA, token-efficiency under 3D-RoPE, and per-skill diagnostics on VideoMME-v2.

\paragraph{Adaptive frame sampling.}
A simpler line of prior work uses learned or heuristic adaptive frame samplers~\citep{tang2025aks,wang2025videoitg} to allocate more frames to motion-intensive segments. The canonical limitation is that the unit of selection is the frame, not the patch: a single frame contributes either every patch or no patches, and the quadratic cost of self-attention is unmitigated for the kept frames. Codec-stream tokenization instead selects at the patch level under a \emph{global} clip-level token budget, so a high-motion frame can contribute a small dense patch cluster and a low-motion frame can contribute its I-frame anchor only.

\subsection{Temporal Grounding with Video LLMs}
\label{subsec:rw_temporal_grounding}

A complementary line of work investigates how video LLMs can localise events in time rather than only describing them. Early approaches augment a frame-sampled video LLM with explicit timestamp tokens or boundary-aware training, including VTimeLLM~\citep{huang2023vtimellm}, TimeChat~\citep{ren2023timechat}, LITA~\citep{huang2024lita}, VTG-LLM~\citep{guo2024vtgllm}, and Grounded-VideoLLM~\citep{wang2024groundedvideollm}, which add per-event captioning supervision or specialised time-token vocabularies on top of the standard uniform-frame interface. TRACE~\citep{guo2024trace} and TimeRefine~\citep{yan2024timerefine} push this further with causal event modelling and iterative boundary refinement. More recent work casts temporal grounding as a reinforcement-fine-tuning problem with verifiable IoU-based rewards: VideoChat-R1~\citep{li2025videochatr1}, Time-R1~\citep{wang2025timer1}, and Video-R1~\citep{feng2025videor1} all post-train a frame-sampled MLLM with GRPO-style updates against grounding rewards. The 2026 cohort extends this trajectory with instance-level grounding (STVG-R1~\citep{zhang2026stvgr1}), curriculum grounding rewards (Video-TwG~\citep{chen2026twg}), event-graph reasoning rewards (GraphThinker~\citep{cheng2026graphthinker}), and verifiable-reward video reasoning on flow-based generative models (Wan-R1~\citep{liu2026wanr1}), alongside broader unified RL post-training recipes for vision-language models~\citep{yan2026sgrpo,xie2026sslr1}. All of these systems retain uniform frame sampling on the input side; their gains come from the supervision objective, not the visual evidence allocation. Codec-stream tokenization is orthogonal: it changes which frames and which patches the model sees in the first place. The two are stackable, and on JumpScore, Charades-STA, ActivityNet, and QVHighlights (\S\ref{sec:codec_analysis}) we show that codec-stream inputs alone, with no temporal-grounding-specific objective, already match or exceed frame-sampled grounding baselines under matched token budgets.

\subsection{Spatial Cognition in Multimodal LLMs}
\label{subsec:rw_spatial}

Spatial cognition has emerged as a distinct evaluation axis where 8B-class video MLLMs underperform their headline results. Cambrian-S~\citep{cambrian-s} formalises ``supersensing'' --- maintaining coherent spatial state across long-horizon video --- as the next bottleneck for video MLLMs and identifies inference-time mechanisms (predictive sensing with surprise-driven memory management) as the most promising direction; we read LLaVA-OneVision-2 as a strong spatially-grounded \emph{base} for that line of work rather than a competing solution to it. Molmo and Pixmo~\citep{molmo} introduced point-based supervision as a structured-output format for static images; Molmo2~\citep{molmo2,clark2026molmo2} extends this to point-based video tracking and spatio-temporal pointing. We adopt the Molmo2 video-tracking and pointing data \emph{as-is} as one ingredient of our Stage-4 spatial supervision (\S\ref{subsec:stage4}); the contribution this paper isolates is codec-stream tokenization, not the point/tracking signal itself. A parallel data-side line of work supplies structured 2D and 3D spatial supervision for MLLMs: VSI-Bench and the ``Thinking in Space'' analysis~\citep{yang2024think}, SpatialBot~\citep{cai2024spatialbot}, SpatialRGPT~\citep{cheng2024spatialrgpt}, EmbodiedScan~\citep{wang2023embodiedscan}, RoboSpatial~\citep{song2024robospatial}, RoboRefer / RefSpatial~\citep{zhou2025roborefer}, SPAR~\citep{zhang2025spar}, MMSI-Bench~\citep{yang2025mmsibench}, and the SG-RLVR recipe of SpaceR~\citep{ouyang2025spacer}. Concurrent 2026 work continues this thread: GR3D~\citep{yuan2026gr3d} attaches geometrically referenced 3D scene representations to MLLMs without training, Spa3R~\citep{jiang2026spa3r} learns view-invariant spatial fields via predictive novel-view feature synthesis, and ``Thinking with Spatial Code''~\citep{chen2026spatialcode} feeds explicit oriented-bounding-box code into the LLM for physical-world video reasoning. New benchmarks tighten the evaluation protocol: ReVSI~\citep{zhang2026revsi} re-annotates VSI-Bench scenes to remove protocol artefacts (we report ReVSI in \S\ref{subsec:main_results}), VAEX-Bench~\citep{bang2026vaexbench} probes abstractive spatiotemporal evidence at object/room/floor-plan level, and SFI-Bench~\citep{zhang2026sfibench} jointly benchmarks spatial and functional intelligence over video. These datasets converge on a similar three-source taxonomy (annotated real videos, simulated trajectories, pseudo-annotated web video) that our \texttt{LLaVA-OneVision-2-Spatial-4M} corpus (\S\ref{subsec:stage4}) follows. The novelty on the data side is not the taxonomy but the decision to pair structured spatial QA with Molmo2-style point/tracking supervision exclusively in Stage 4 of the recipe; \S\ref{subsec:stage4} explains why this stage ordering matters empirically.

\subsection{Referring Video Object Segmentation with Multimodal LLMs}
\label{subsec:rw_rvos}

Referring video object segmentation (R-VOS) on MeViS~\citep{ding2023mevis} and ReVOS asks the model to follow a single physical reference across many frames given a language query, and is the closest public proxy for the cross-frame coherence that robotics deployment needs. Recent MLLM-based R-VOS systems pair an LLM with a promptable segmentation backbone, typically SAM 2~\citep{sam2}: VideoLISA~\citep{bai2024videolisa} introduces a single \texttt{<TRK>} token under a sparse-dense sampling regime, Sa2VA~\citep{yuan2025sa2va} marries SAM 2 with LLaVA-OneVision for unified image / video grounding, VideoGLaMM~\citep{munasinghe2024videoglamm} extends GLaMM~\citep{rasheed2023glamm} to pixel-level video grounding, Vitron~\citep{fei2024vitron} unifies twelve image/video tasks under one pixel-level LLM, and VideoMolmo~\citep{rasheed2025videomolmo} couples Molmo-style point supervision with bidirectional SAM 2 propagation. Concurrent 2026 work pushes toward agentic and tracking-aware R-VOS: AgentRVOS~\citep{jin2026agentrvos} first generates SAM 3 object tracks and then asks the LLM to reason over them for zero-shot R-VOS, while SPARROW~\citep{alansari2026sparrow} learns dual box/segmentation prompts with tracked features for spatially precise, temporally consistent pixel grounding.
\section{Conclusion}
\label{sec:conclusion}

We presented \textbf{LLaVA-OneVision-2} as a codec-aligned long-video MLLM that moves video observation beyond frame-centric sampling toward stream-aware perceptual evidence allocation. Built on a native-resolution OneVision-Encoder, the model treats compressed video as a continuous bit-cost stream: bit-cost dynamics determine adaptive temporal groups, while motion-residual evidence is condensed into compact visual canvases and processed through a unified group-visible attention interface. Together with a progressive open training stack of approximately \(8\)M re-captioned video samples, a \(4\)M-sample 2D\&3D spatial corpus, and the JumpScore temporal-localization benchmark, LLaVA-OneVision-2 demonstrates strong gains across video, spatial, temporal grounding, and tracking evaluations. More importantly, our analysis shows that codec-stream tokenization is not merely a token-reduction mechanism, but a perceptual allocation principle: it expands event-level coverage for long-video reasoning while retaining frame sampling as a complementary path for detail-sensitive perception. Looking forward, we will extend this codec-aligned paradigm toward streaming perception and hours-scale or longer codec-context modeling~\citep{guan2026vst,shen2026simplestream,chen2026flexmem}, where visual evidence must be continuously updated, compressed, and retrieved over ultra-long temporal horizons.         
\section{Contributors}

\begin{minipage}[t]{0.48\textwidth}
\raggedright
\textbf{Contributors} \\ 
{\color{gray}\small core contributors are in bold}

\begin{itemize}[noitemsep,topsep=0pt,leftmargin=*]

\item \textbf{Xiang An}
\item \textbf{Yin Xie}
\item \textbf{Feilong Tang}
\item \textbf{Yunyao Yan}
\item \textbf{Huajie Tan}
\item \textbf{Didi Zhu}
\item \textbf{Changrui Chen}
\item \textbf{Xiuwei Zhao}
\item Bin Qin
\item Kaicheng Yang
\item Yifei Shen
\item Yuanhan Zhang
\item Kaichen Zhang
\item Wenkang Zhang
\item Zheng Cheng
\item Manyuan Zhang
\item Chunsheng Wu
\item Chunjiang Ge
\item Zimin Ran
\item Dehua Song
\item Chunyuan Li
\item Shikun Feng
\item Ming Hu
\item Zhangquan Chen
\item Junbo Niu

\end{itemize}
\end{minipage}\hfill
\begin{minipage}[t]{0.48\textwidth}
\raggedright
\textbf{Project Leaders}

\begin{itemize}[noitemsep,topsep=0pt,leftmargin=*]
\item Bo Li
\item Ziyong Feng
\item Ziwei Liu
\item Zongyuan Ge
\item Jiankang Deng
\end{itemize}
\end{minipage}

\clearpage

\section{Details}
\label{sec:implementation}

\subsection{Codec-Stream versus Uniform Frame Sampling}
\label{app:codec_vs_frame_values}

Figure~\ref{fig:codec_vs_frame} compares codec-stream inputs with uniformly sampled RGB frames under matched nominal frame budgets. 
To make the scaling curves auditable, we report the exact numerical values used in the figure in Tables~\ref{tab:app_codec_vs_frame_videoqa} and~\ref{tab:app_codec_vs_frame_grounding}. 
``Uniform'' denotes standard uniform frame sampling, and ``Codec'' denotes the codec-stream input representation.

\begin{table*}[h]
\centering
\small
\setlength{\tabcolsep}{5pt}
\renewcommand{\arraystretch}{1.08}
\caption{\textbf{Numerical values for the video QA and JumpScore curves in Figure~\ref{fig:codec_vs_frame}.}
Each benchmark is reported under matched nominal frame budgets.}
\label{tab:app_codec_vs_frame_videoqa}
\adjustbox{max width=\textwidth}{
\begin{tabular}{c|cc|cc|cc|cc}
\toprule
\multirow{2}{*}{\textbf{Budget}}
& \multicolumn{2}{c|}{\textbf{VideoMME-L-sub}}
& \multicolumn{2}{c|}{\textbf{LVBench}}
& \multicolumn{2}{c|}{\textbf{VideoEval-Pro}}
& \multicolumn{2}{c}{\textbf{JumpScore}} \\
\cmidrule(lr){2-3}
\cmidrule(lr){4-5}
\cmidrule(lr){6-7}
\cmidrule(lr){8-9}
& \textbf{Uniform} & \textbf{Codec}
& \textbf{Uniform} & \textbf{Codec}
& \textbf{Uniform} & \textbf{Codec}
& \textbf{Uniform} & \textbf{Codec} \\
\midrule

4
& -- & --
& -- & --
& -- & --
& 32.5 & 39.4 \\

8
& 55.1 & 56.6
& -- & --
& 43.1 & 46.4
& 35.2 & 40.2 \\

16
& 57.7 & 60.3
& 38.9 & 40.9
& 44.8 & 48.3
& 36.7 & 46.9 \\

32
& 62.0 & 61.4
& 41.6 & 42.9
& 48.6 & 51.6
& 37.6 & 58.3 \\

64
& 64.8 & 67.4
& 45.6 & 47.1
& 52.3 & 53.9
& 39.9 & 71.3 \\

128
& 67.1 & 67.0
& 47.7 & 49.5
& 54.0 & 55.8
& 45.4 & 74.9 \\

\bottomrule
\end{tabular}}
\end{table*}

\begin{table*}[h]
\centering
\small
\setlength{\tabcolsep}{6pt}
\renewcommand{\arraystretch}{1.08}
\caption{\textbf{Numerical values for the temporal grounding curves in Figure~\ref{fig:codec_vs_frame}.}
Each benchmark is reported under matched nominal frame budgets.}
\label{tab:app_codec_vs_frame_grounding}
\adjustbox{max width=\textwidth}{
\begin{tabular}{c|cc|cc|cc}
\toprule
\multirow{2}{*}{\textbf{Budget}}
& \multicolumn{2}{c|}{\textbf{QVHighlights}}
& \multicolumn{2}{c|}{\textbf{Charades-STA}}
& \multicolumn{2}{c}{\textbf{ActivityNet Captions}} \\
\cmidrule(lr){2-3}
\cmidrule(lr){4-5}
\cmidrule(lr){6-7}
& \textbf{Uniform} & \textbf{Codec}
& \textbf{Uniform} & \textbf{Codec}
& \textbf{Uniform} & \textbf{Codec} \\
\midrule

4
& 12.3 & 27.7
& 17.4 & 42.4
& 12.0 & 23.1 \\

8
& 23.3 & 40.4
& 37.1 & 48.2
& 18.5 & 31.8 \\

16
& 24.9 & 51.5
& 48.1 & 51.0
& 28.1 & 40.6 \\

32
& 53.2 & 59.2
& 53.1 & 49.9
& 39.9 & 47.5 \\

64
& 61.7 & 63.5
& 53.5 & 50.1
& 48.9 & 51.2 \\

\bottomrule
\end{tabular}}
\end{table*}

Across the general long-video QA benchmarks, codec-stream inputs provide modest but mostly positive gains over uniform frame sampling. 
The average absolute improvements are 1.2 points on VideoMME-L-sub, 1.7 points on LVBench, and 2.6 points on VideoEval-Pro. 
The gains are not strictly monotonic at every budget: for example, VideoMME-L-sub shows small drops at 32 and 128 frames. 
Nevertheless, the overall pattern suggests that codec-stream tokenization can preserve useful temporal evidence without increasing the nominal frame budget.

The advantage becomes larger on motion-sensitive temporal localization tasks. 
On JumpScore, codec-stream improves the average score from 37.9 to 55.2 across the six evaluated budgets, corresponding to an average absolute gain of 17.3 points. 
At the 4-frame budget, the score improves from 32.5 to 39.4, which corresponds to 6.9 absolute points and 21.2\% relative improvement over uniform frame sampling. 
The gains further increase at larger budgets, reaching 20.7, 31.4, and 29.5 points at 32, 64, and 128 frames, respectively.

For temporal grounding, codec-stream is most beneficial under low- and medium-frame budgets. 
On QVHighlights, the gains are 15.4, 17.1, and 26.6 points at 4, 8, and 16 frames, respectively. 
On Charades-STA, codec-stream substantially improves the 4-frame setting from 17.4 to 42.4, while uniform sampling becomes competitive at higher budgets. 
On ActivityNet Captions, codec-stream improves over uniform sampling at all evaluated budgets, with gains ranging from 2.3 to 13.3 points. 
Overall, the numerical values behind Figure~\ref{fig:codec_vs_frame} show that codec-stream inputs provide the largest benefit when temporally sparse, motion-relevant evidence must be recovered from a limited visual budget.

\subsection{Video Tracking}

\paragraph{Tracking protocol.}
Across all benchmarks, segmentation metrics are computed at the original video frame rate, whereas point-based metrics are evaluated at 1 fps and marked as correct when a predicted point falls inside the ground-truth mask. 
For segmentation-capable baselines, we evaluate the masks produced by the corresponding open segmentation models and report their mask-level quality. 
When a model can produce discrete points for the referred object, such as VLM-style point outputs, we additionally evaluate its point-based localization accuracy. 
LLaVA-OneVision-2-8B does not introduce a new segmentation head. 
Instead, it predicts discrete point tracks with explicit object IDs, and these ID-associated points are directly used as SAM~2 point prompts to obtain per-frame segmentation masks. 
The reported mask metrics for our model therefore evaluate the quality of ID-consistent point tracking together with the point-to-mask conversion, rather than a separately trained segmentation decoder.

\trackingbenchmarktable

\paragraph{Results.}
Table~\ref{tab:tracking_main} shows that LLaVA-OneVision-2-8B achieves the best overall tracking score, improving over Qwen3-VL-8B by \textbf{+10.2} absolute points (41.0 vs.\ 30.8). 
The gain is consistent on the mask-overlap metric J\&F across all four evaluation splits: +17.4 on DAVIS (58.7 vs.\ 41.3), +17.3 on MeViS-U (45.7 vs.\ 28.4), +20.4 on REVOS-Referring (58.2 vs.\ 37.8), and +7.3 on REVOS-Reasoning (29.2 vs.\ 21.9). 
These results support our hypothesis that codec-stream dense observation is useful for cross-frame correspondence: at matched token budgets, retaining I-frame patches densely while sparsifying P-frames toward motion-relevant regions provides more temporally aligned visual evidence per token. 
The Stage-4 point-based tracking supervision from Molmo2-VideoTrack provides a complementary mechanism, teaching the model to maintain a single physical reference across frames instead of re-grounding the object from text independently at every frame. 
Although Qwen3-VL-8B remains stronger on HOTA for MeViS-U and REVOS-Reasoning, LLaVA-OneVision-2-8B obtains higher J\&F on every split, indicating better mask overlap after point-to-mask conversion.

\newpage 
\section{Case Study}
\label{sec:Cases}


\subsection{Temporal Grounding on TimeLens-Bench}

\begin{figure*}[!htbp]
    \centering
    \includegraphics[width=\linewidth]{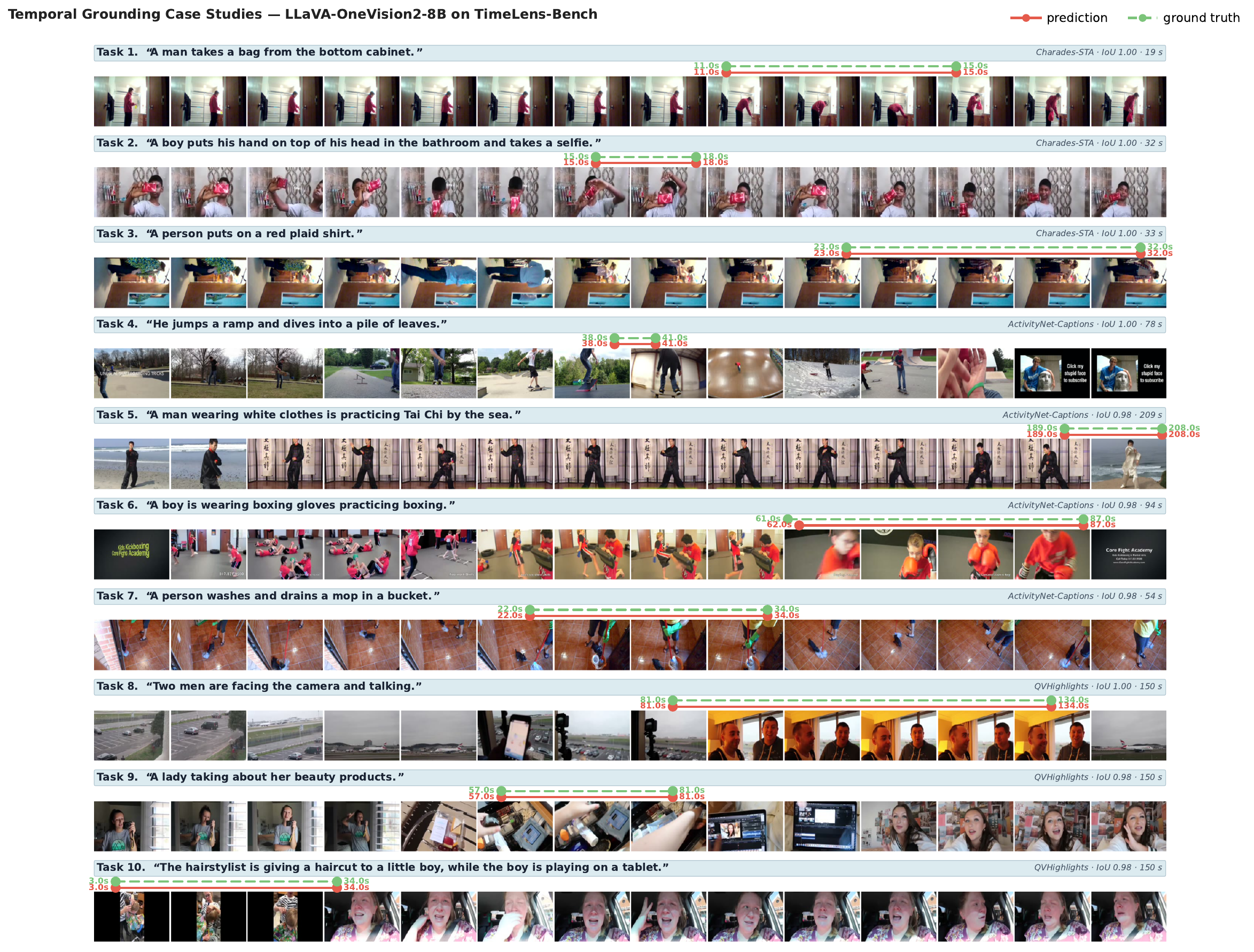}
    \caption{\textbf{Temporal grounding on TimeLens-Bench.} Ten cases from Charades-STA, ActivityNet-Captions, and QVHighlights; per row, mean IoU over five runs.}
    \label{fig:case_temporal_grounding}
\end{figure*}

Figure~\ref{fig:case_temporal_grounding} reports ten temporal-grounding cases from Charades-STA, ActivityNet-Captions, and QVHighlights; per-row metric is mean IoU over five runs.

\clearpage


\subsection{Referring Video Object Segmentation and Tracking on ReasonVOS}

\begin{figure*}[!htbp]
    \centering
    \includegraphics[width=0.95\linewidth]{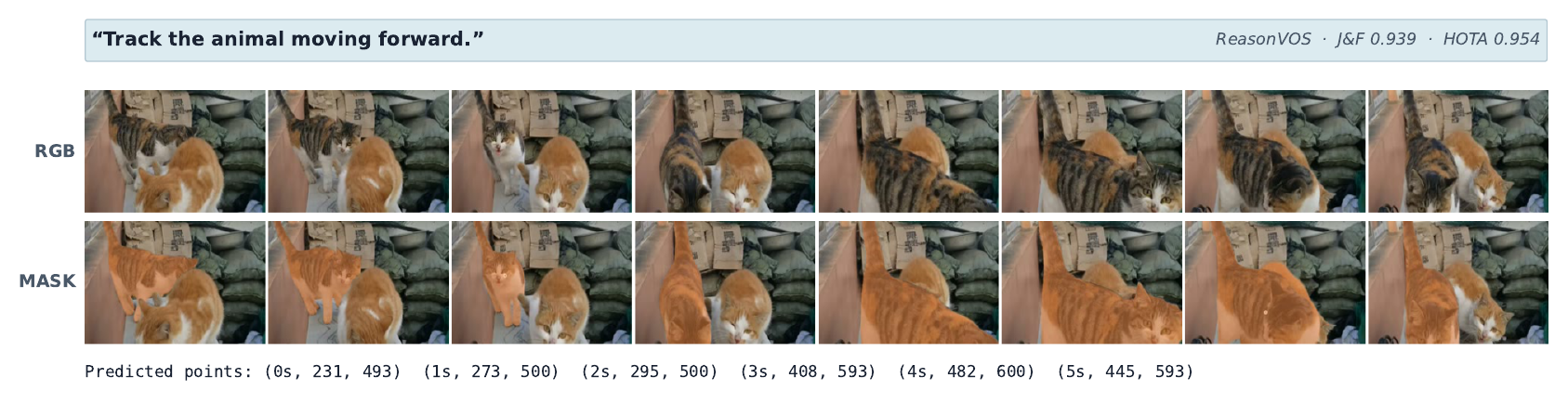}
    \caption{\textbf{R-VOS on ReasonVOS --- ``Track the animal moving forward''.} The two strips show eight evenly-sampled frames of the input RGB (top) and the corresponding SAM2-derived dense mask (bottom) for a 36-frame clip; the per-frame $(x, y)$ tracking points emitted by our model and used as SAM2 prompts are listed below the strips.}
    \label{fig:case_rvos_reasonvos}
\end{figure*}

ReasonVOS evaluates referring video object segmentation given a free-form referring expression and a video. Our model emits a per-frame $(x, y)$ tracking point for the referred object rather than a dense mask directly; the dense mask shown in the figure is produced by feeding these per-frame points to SAM2~\citep{sam2} downstream as prompts. Figure~\ref{fig:case_rvos_reasonvos} shows a 36-frame ``Track the animal moving forward'' case in which a cat moves diagonally across the scene; the point sequence follows the cat through pose change and partial occlusion, yielding $\mathcal{J}\&\mathcal{F}=0.939$ and HOTA $=0.954$ on the SAM2 mask.

\subsection{Referring Video Object Segmentation and Tracking on Ref-DAVIS17}

\begin{figure*}[!htbp]
    \centering
    \includegraphics[width=0.95\linewidth]{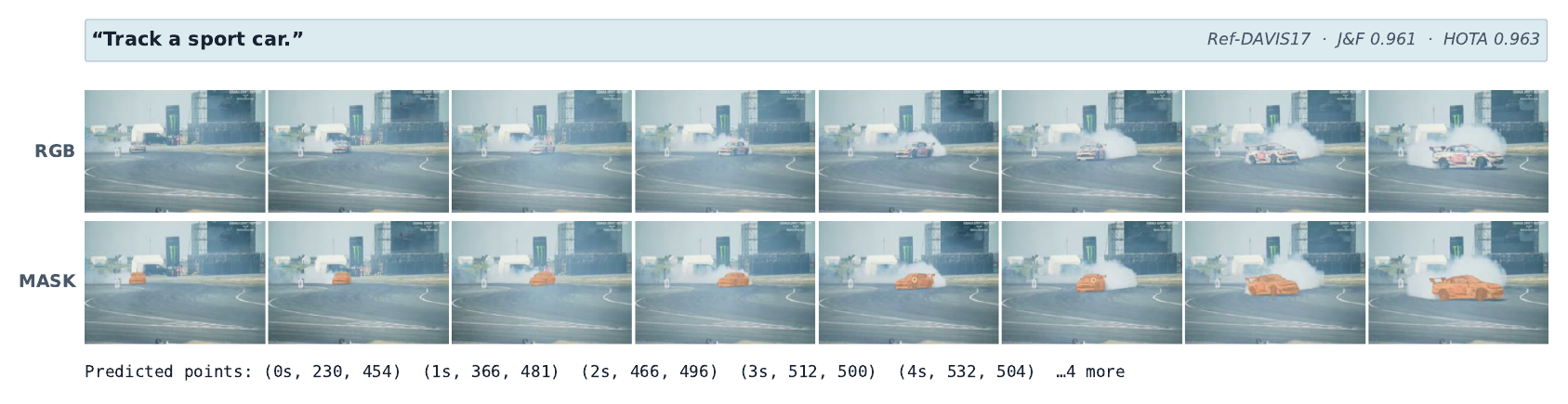}
    \caption{\textbf{R-VOS on Ref-DAVIS17 --- ``Track a sport car''.} Same two-strip layout as Figure~\ref{fig:case_rvos_reasonvos}: the top strip is the input RGB and the bottom strip is the SAM2-derived dense mask, with the model's per-frame $(x, y)$ points printed below.}
    \label{fig:case_rvos_davis17}
\end{figure*}

Ref-DAVIS17 evaluates the same referring video task on a different visual domain that emphasises high-motion outdoor footage. The pipeline is identical to the ReasonVOS case above: per-frame tracking points are emitted by our model and the dense mask is recovered by SAM2~\citep{sam2} downstream. Figure~\ref{fig:case_rvos_davis17} shows a 52-frame Drift-Chicane ``Track a sport car'' case where the point sequence stays aligned through tire smoke, motion blur, and viewpoint change, yielding $\mathcal{J}\&\mathcal{F}=0.961$ and HOTA $=0.963$ on the SAM2 mask.

\clearpage


\subsection{Real-World Robot Manipulation}

We deploy the model on two tabletop manipulation tasks (apple-to-plate, left; bread-to-oven, right) and re-query it online at three moments during execution. At each query the model receives only the natural-language instruction and the current RGB observation, and returns an image-space list of $(x, y, z)$ waypoints that a downstream IK module follows; the waypoint count contracts as the gripper approaches the target, and the trajectory updates whenever the scene state changes.

\begin{figure*}[!htbp]
\centering
\definecolor{taskbarcolor}{RGB}{220, 235, 240}

\begin{minipage}{0.95\textwidth}
\centering
\begin{minipage}[t]{0.495\linewidth}
\colorbox{taskbarcolor}{\parbox{0.985\linewidth}{\centering\scriptsize\textbf{Task 1:} ``Put the apple on the green plate placed on the table.''}}

\vspace{3pt}

\begin{minipage}[t]{0.325\linewidth}\centering
  \includegraphics[width=\linewidth]{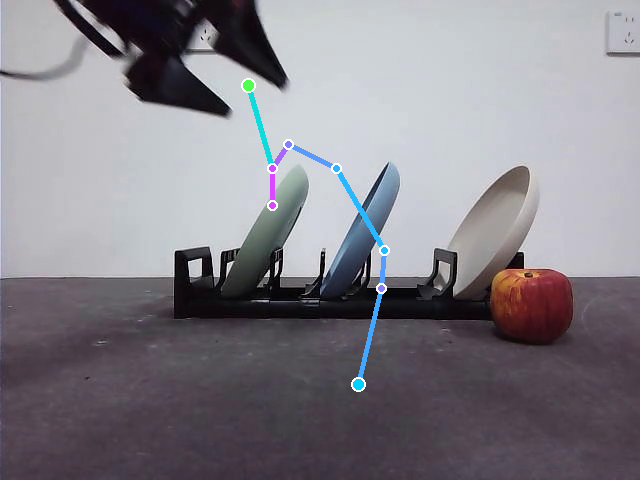}\\[1pt]
  {\scriptsize $t{=}0\,$s\,\textcolor{gray}{\,(9 wpts)}}
\end{minipage}\hfill
\begin{minipage}[t]{0.325\linewidth}\centering
  \includegraphics[width=\linewidth]{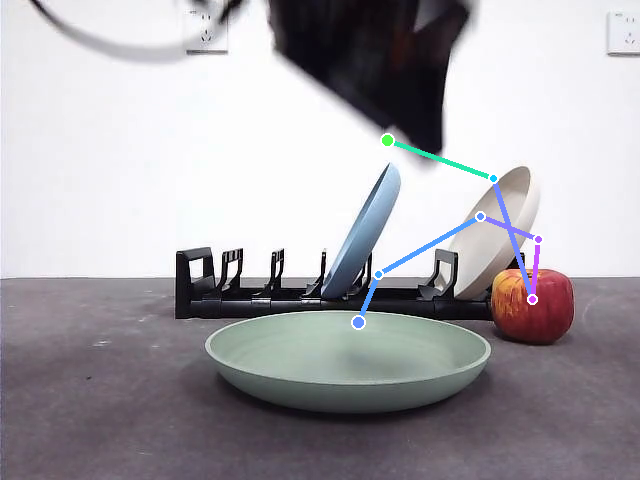}\\[1pt]
  {\scriptsize $t{=}10\,$s\,\textcolor{gray}{\,(7 wpts)}}
\end{minipage}\hfill
\begin{minipage}[t]{0.325\linewidth}\centering
  \includegraphics[width=\linewidth]{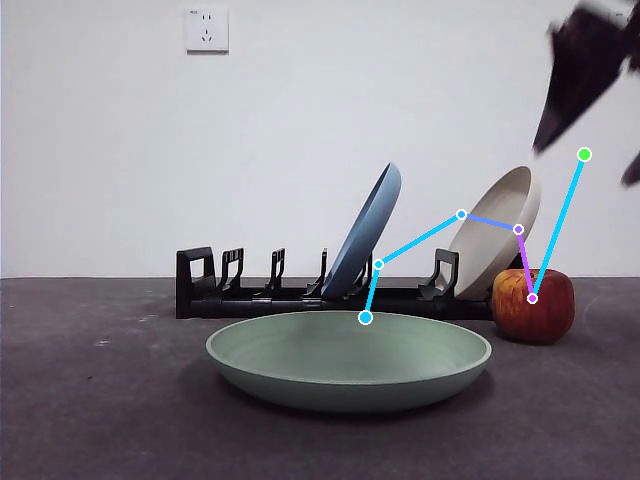}\\[1pt]
  {\scriptsize $t{=}11\,$s\,\textcolor{gray}{\,(6 wpts)}}
\end{minipage}
\end{minipage}\hfill
\begin{minipage}[t]{0.495\linewidth}
\colorbox{taskbarcolor}{\parbox{0.985\linewidth}{\centering\scriptsize\textbf{Task 2:} ``Put the bread into the oven.''}}

\vspace{3pt}

\begin{minipage}[t]{0.325\linewidth}\centering
  \includegraphics[width=\linewidth]{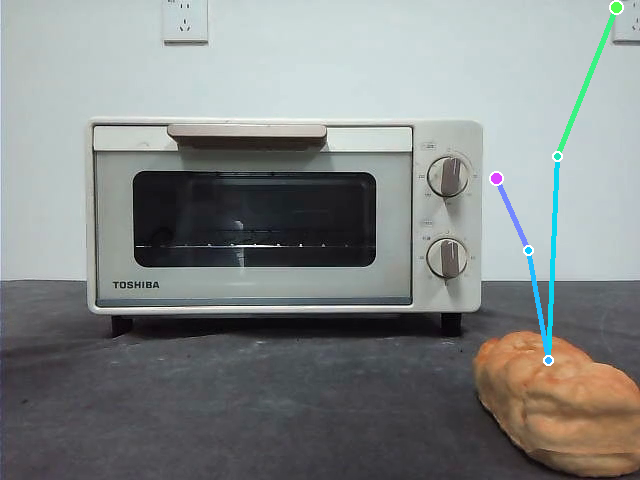}\\[1pt]
  {\scriptsize $t{=}0\,$s\,\textcolor{gray}{\,(5 wpts)}}
\end{minipage}\hfill
\begin{minipage}[t]{0.325\linewidth}\centering
  \includegraphics[width=\linewidth]{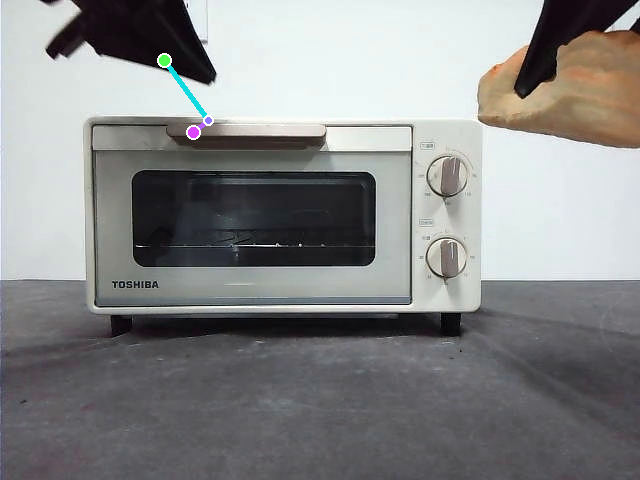}\\[1pt]
  {\scriptsize $t{=}7\,$s\,\textcolor{gray}{\,(3 wpts)}}
\end{minipage}\hfill
\begin{minipage}[t]{0.325\linewidth}\centering
  \includegraphics[width=\linewidth]{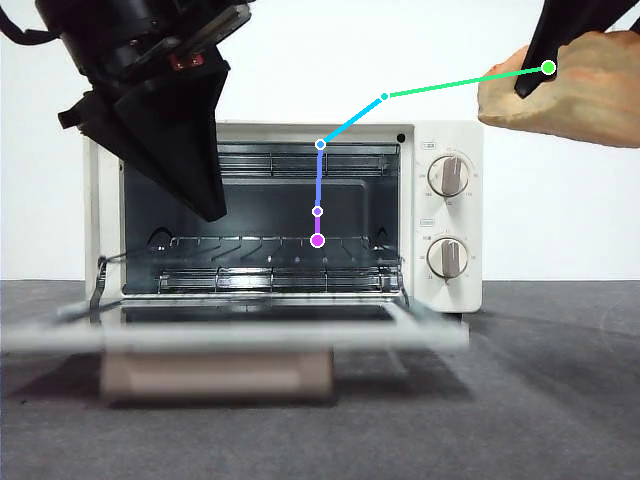}\\[1pt]
  {\scriptsize $t{=}13\,$s\,\textcolor{gray}{\,(5 wpts)}}
\end{minipage}
\end{minipage}
\end{minipage}

\caption{\textbf{Real-world trajectory predictions on a robot manipulation setup.} Cyan polyline $=$ trajectory order; magenta dots $=$ waypoints; green dot $=$ start; $z$ is a normalised depth.}
\label{fig:real_robot_demo}
\end{figure*}

\subsection{JumpScore Validation}

\begin{figure*}[!htbp]
    \centering
    \includegraphics[width=0.80\linewidth]{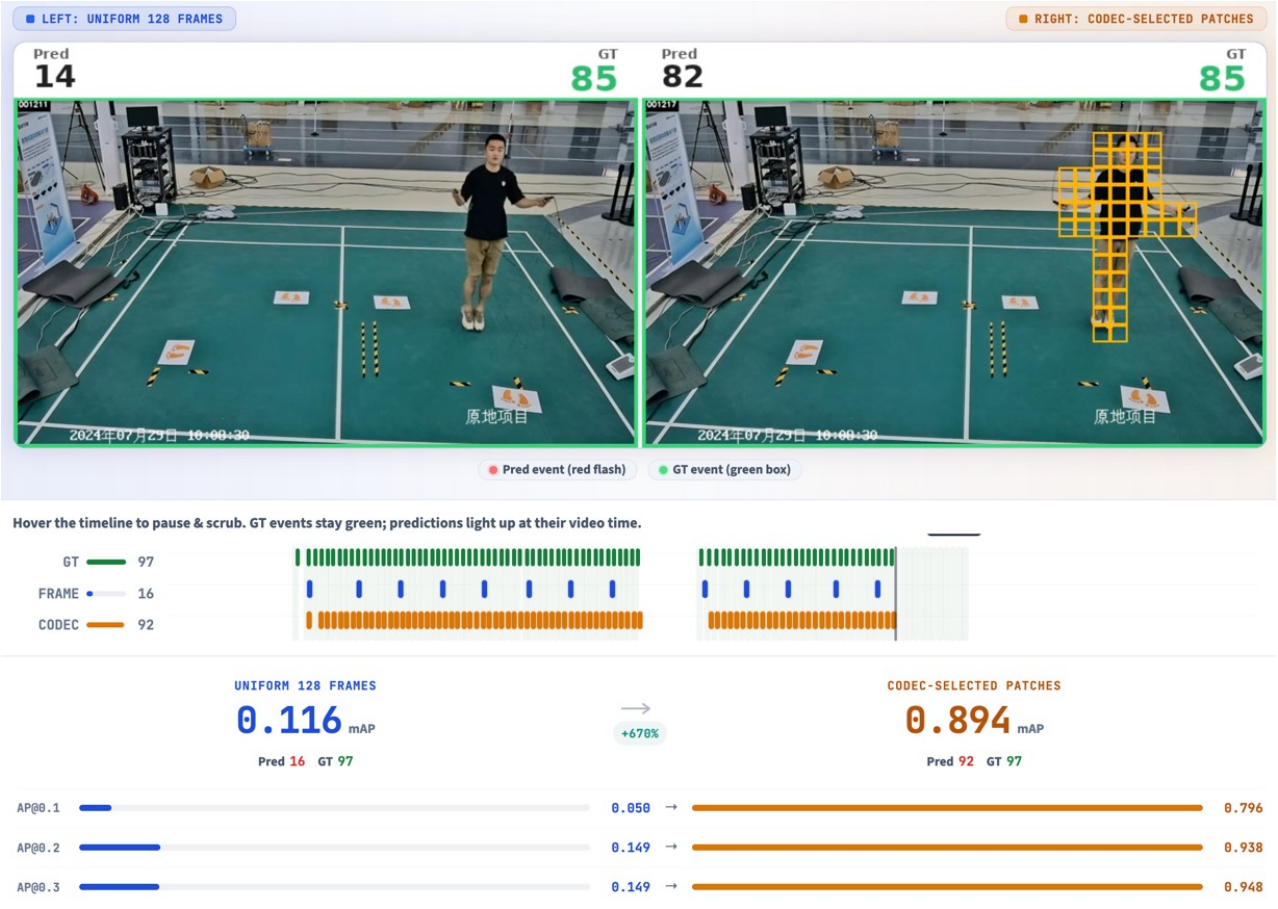}
    \caption{\textbf{JumpScore validation: uniform vs.\ codec-stream sampling on a single 85-cycle clip.} At matched visual-token budget, codec-stream sampling attributes 82 of 85 cycle starts (mIoU $0.894$) versus 14 of 85 for uniform 128-frame sampling (mIoU $0.116$); each predicted cycle start is drawn green when it lands within $0.1$\,s of a ground-truth start and red otherwise.}
    \label{fig:case_jumpscore}
\end{figure*}

Figure~\ref{fig:case_jumpscore} compares uniform 128-frame sampling and codec-stream sampling on a single JumpScore clip with 85 ground-truth cycle starts at matched visual-token budget. Uniform sampling attributes 14 of 85 cycles correctly (mean IoU $0.116$), while codec-stream sampling attributes 82 of 85 (mean IoU $0.894$). The improvement reflects codec sampling's ability to concentrate visual tokens on motion-residual regions, exactly where cycle boundaries live in densely repeated jump-rope motion.

\clearpage


\subsection{2D Spatial Grounding}

\begin{figure*}[!htbp]
    \centering
    \includegraphics[width=\linewidth]{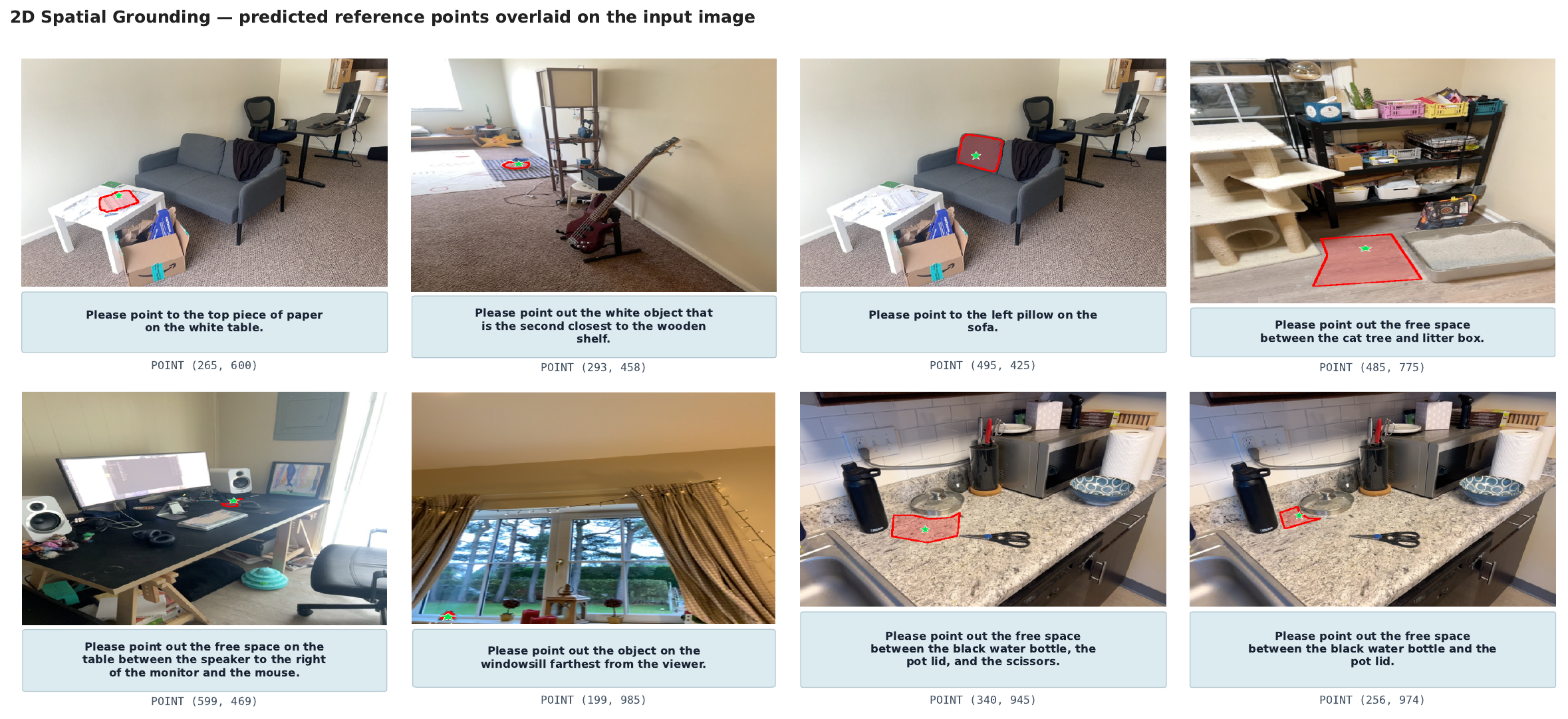}
    \caption{\textbf{2D spatial grounding.} Eight examples; predicted point overlaid in red.}
    \label{fig:case_spatial_2d}
\end{figure*}

Figure~\ref{fig:case_spatial_2d} reports eight 2D pointing cases covering object references, relational queries, and free-space queries; the model emits an $(x, y)$ pixel coordinate (overlaid in red).

\subsection{3D Spatial Grounding}

\begin{figure*}[!htbp]
    \centering
    \includegraphics[width=\linewidth]{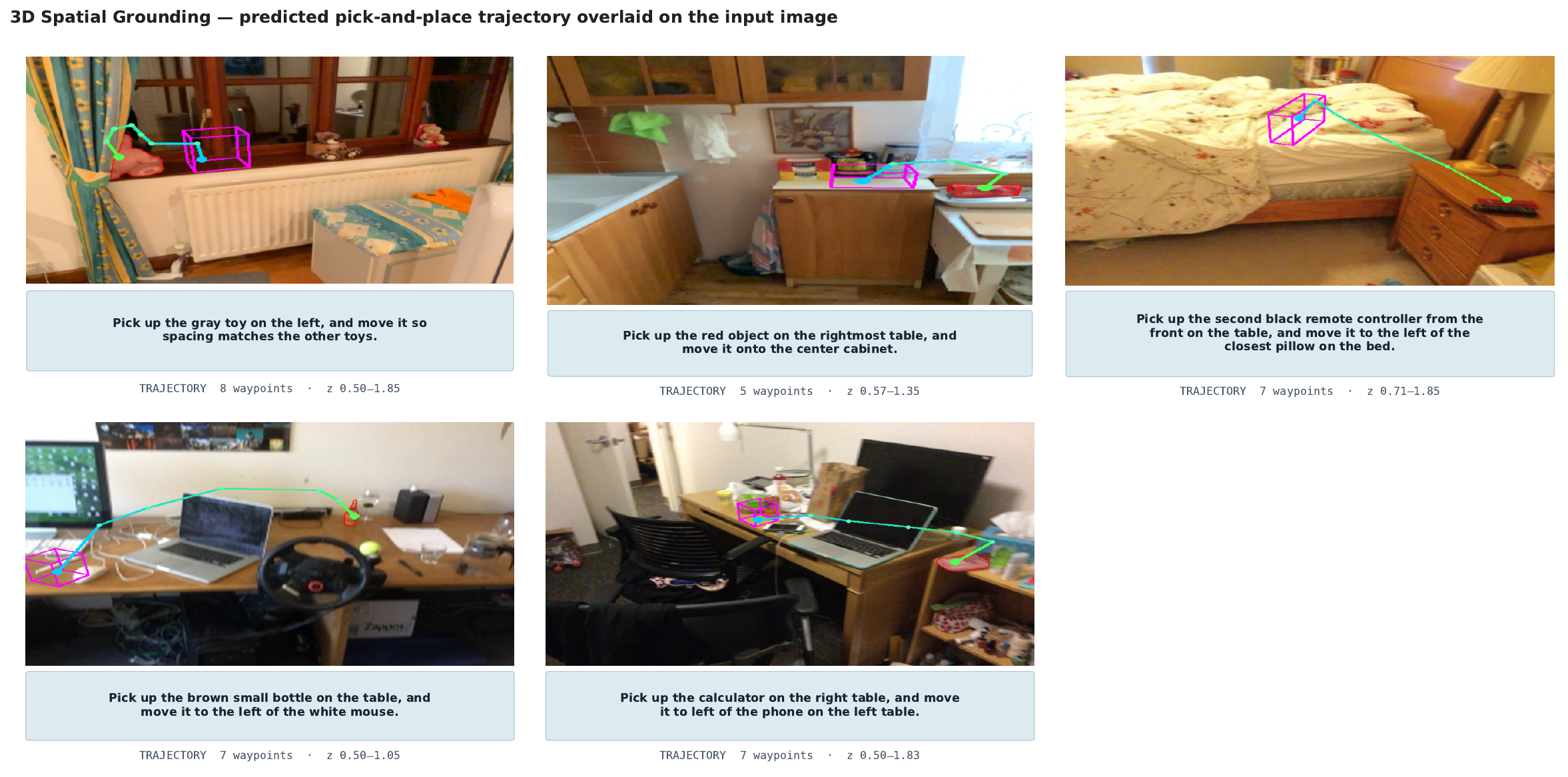}
    \caption{\textbf{3D spatial grounding.} Five pick-and-place examples; predicted 3D trajectory overlaid.}
    \label{fig:case_spatial_3d}
\end{figure*}

Figure~\ref{fig:case_spatial_3d} reports five 3D pick-and-place cases (``pick up X, and move it to Y''); the model emits a continuous 3D trajectory in image space with relative depth.

\clearpage

{
    \small
    \bibliographystyle{ieeenat_fullname}
    \bibliography{main}
}

\end{document}